\documentclass[10pt,preprint,review]{elsarticle}
\usepackage{amsmath,amsfonts}
\usepackage{algorithmic}
\usepackage{array}
\usepackage{textcomp}
\usepackage{stfloats}
\usepackage{url}
\usepackage{verbatim}
\usepackage{graphicx}
\usepackage{subcaption}
\usepackage{bm}
\usepackage{float} 
\usepackage{booktabs}
\usepackage{array, caption, threeparttable}
\usepackage[font=small,labelfont=bf,labelsep=none]{caption}
\usepackage{balance}
\usepackage{algorithm}
\usepackage{natbib}
\setcitestyle{numbers,square}
\usepackage{color}
\usepackage{stfloats}
\makeatletter

\usepackage{hyperref}
\journal{}
\makeatother

\begin{document}
	
\begin{frontmatter}
	
\title{Symmetrization Weighted Binary Cross-Entropy: Modeling Perceptual Asymmetry for Human-Consistent Neural Edge Detection}

\author{Hao Shu\corref{cor1}}
\ead{Hao_B_Shu@163.com}
\cortext[cor1]{}

\affiliation{organization={Sun-Yat-Sen University},
	city={Shenzhen},
	country={China}}
\affiliation{organization={Shenzhen University},
	city={Shenzhen},
	country={China}}

\begin{abstract}
Edge detection (ED) is a fundamental perceptual process in computer vision, forming the structural basis for high-level reasoning tasks such as segmentation, recognition, and scene understanding. Despite substantial progress achieved by deep neural networks, most ED models attain high numerical accuracy but fail to produce visually sharp and perceptually consistent edges, thereby limiting their reliability in intelligent vision systems. To address this issue, this study introduces the \textit{Symmetrization Weighted Binary Cross-Entropy (SWBCE)} loss, a perception-inspired formulation that extends the conventional WBCE by incorporating prediction-guided symmetry. SWBCE explicitly models the perceptual asymmetry in human edge recognition, wherein edge decisions require stronger evidence than non-edge ones, aligning the optimization process with human perceptual discrimination. The resulting symmetric learning mechanism jointly enhances edge recall and suppresses false positives, achieving a superior balance between quantitative accuracy and perceptual fidelity. Extensive experiments across multiple benchmark datasets and representative ED architectures demonstrate that SWBCE can outperform existing loss functions in both numerical evaluation and visual quality. Particularly with the HED-EES model, the SSIM can be improved by about 15\% on BRIND, and in all experiments, training by SWBCE consistently obtains the best perceptual results. Beyond edge detection, the proposed perceptual loss offers a generalizable optimization principle for soft computing and neural learning systems, particularly in scenarios where asymmetric perceptual reasoning plays a critical role.
\end{abstract}

\begin{highlights}
   \item Introduce SWBCE, a perception-inspired loss function enhancing edge detection.
    \item Model human-like perceptual asymmetry through prediction-guided weighting.
    \item Achieve a balance between perceptual fidelity and quantitative accuracy without post-processing.
    \item Demonstrate consistent gains across diverse datasets and architectures.
    \item Enhance the reliability of intelligent vision systems through perceptual optimization.
\end{highlights}


\end{frontmatter}

\section{Introduction}

The rapid advancement of computational intelligence has established machine learning as a transformative paradigm for modeling and interpreting complex real-world phenomena. A broad spectrum of methodologies—from classical statistical models\cite{CV1995Support,T1996Regression,Q1996Learning,DL1977Maximum,GG2022Principal} to modern data-driven neural networks\cite{KS2012Imagenet,GD2024Mamba}—has been widely applied across multiple levels of tasks, ranging from fundamental applications such as data classification\cite{RL2014Clustering,S2025SDC}, natural language processing\cite{VS2017Attention}, and computer vision\cite{JK2025Generative}, to high-level system integrations including autonomous driving\cite{SL2025Sparsedrive} and industrial automation\cite{BA2023Industrial}. Despite these achievements, enabling intelligent systems to effectively adapt to the intrinsic complexity and uncertainty of real-world environments remains a long-standing challenge.

Considerable research efforts have been devoted to addressing this challenge. For example, Gaussian Process Regression\cite{WR1995Gaussian,Ma2025Probabilistic} shifts modeling from deterministic predictions to probabilistic distributions, offering a principled framework for uncertainty quantification. Graph-based methods\cite{KF2009Probabilistic,BH2018Relational,HR2025Verilogcoder} capture structured dependencies among entities, thereby enhancing relational coherence during inference. In addition, ensemble methods\cite{Z2025Ensemble,LX2025Multi} combine multiple algorithms to improve overall robustness and generalization. Nevertheless, faithfully capturing the rich and nuanced uncertainty of natural processes—particularly the complex cognitive mechanisms underlying human perception—continues to represent a fundamental and unresolved frontier.

In this paper, we focus on edge detection (ED), which represents a canonical instance of modeling such intricate cognitive processes and serves as a foundational cornerstone in computer vision. Within the paradigm of soft computing, ED is not merely a pixel-level classification task but a sophisticated form of perceptual reasoning. It aims to extract meaningful structural cues from uncertain and noisy visual environments, providing essential perceptual foundations for high-level tasks such as image inpainting \cite{NN2019EdgeConnect}, object detection \cite{ZD2007An}, and semantic segmentation \cite{MR2011Edge}.

Over the past decade, Convolutional Neural Networks (CNNs) have substantially advanced ED by learning hierarchical feature representations directly from raw data. Representative architectures such as Holistically-Nested Edge Detection (HED) \cite{XT2015Holistically}, Bi-Directional Cascade Network (BDCN) \cite{HZ2022BDCN}, and Dense Extreme Inception Network (Dexi) \cite{SS2023Dense} have achieved competitive quantitative performance. More recently, frameworks like the Extractor-Selector (E-S) \cite{S2025Boost} have further improved edge localization by decoupling the task into two stages: the extraction of rich feature representation followed by a selective refinement stage to suppress noise and sharpen boundaries.

Despite these algorithmic advances, most existing methods remain constrained by a critical trade-off between numerical accuracy and perceptual fidelity. Their predictions, while achieving high statistical scores, frequently appear visually inconsistent or perceptually unfriendly to human observers. This gap can lead to severe cascading consequences when ED is integrated into downstream intelligent systems. As a front-end cue-extraction module, the perceptual degradation may propagate through the network hierarchy and be amplified by subsequent processing stages. In image inpainting, for instance, perceptually flawed edges—despite their high statistical scores—can result in jarring structural artifacts that render the restored regions visually "fake" and contextually incoherent. Such misalignment highlights the need for moving beyond simple statistical balance to prioritize human-like perceptual alignment.

From the perspective of neural learning theory, the perceptual limitation largely stems from loss function design. Conventional losses such as Binary Cross-Entropy (BCE) and its weighted variant (WBCE) primarily address class imbalance from a statistical standpoint but overlook a crucial property of human perception: cognitive asymmetry. Humans tend to confirm edges only when strong evidence is available, while non-edges are generally assumed by default. This inherent asymmetry suggests that edge prediction should be cautious and evidence-driven, rather than statistically balanced alone. Hence, a perception-aware formulation that captures this asymmetry could better align optimization with human-like reasoning in vision models. However, this perceptual asymmetry has rarely been explicitly formulated in loss function design, leaving a gap between perceptual understanding and neural optimization.

Motivated by this observation, we propose a novel loss function termed Symmetrization Weighted Binary Cross-Entropy (SWBCE). SWBCE extends WBCE by introducing prediction-guided weighting, thereby establishing a symmetric learning mechanism between label-based and prediction-based supervision. As such, it explicitly models perceptual asymmetry, enforcing a rigorous evidence-driven criterion for edge confirmation while preserving the class-balancing efficacy of the original WBCE. As a result, this formulation enhances edge reliability and visual sharpness without compromising quantitative performance. Extensive experiments across multiple datasets and architectures demonstrate that SWBCE consistently bridges the gap between perceptual quality and numerical accuracy, yielding edge predictions that are both perceptually human-aligned and quantitatively robust. Particularly with the HED-EES model, the SSIM can be improved by about 15\% on BRIND, and in all experiments, training by SWBCE consistently obtains the best perceptual results.

The main contributions of this work are summarized as follows:
\begin{enumerate}
    \item We analyze the limitations of conventional WBCE losses and highlight the significance of incorporating perceptual asymmetry into loss design for edge detection.
    \item We propose the SWBCE loss, a symmetric, perception-aware formulation that unifies label- and prediction-guided supervision to balance edge recall and false-positive suppression.
    \item We define a new evaluation criterion termed \textbf{Edge Ratio}, which measures the proportion of correct edge predictions relative to the total prediction set from a continuous perspective, assessing the edge precision beyond binarization. It can be useful when binarization is not allowed.
    \item We conduct comprehensive experiments across multiple public benchmarks, validating the effectiveness, perceptual consistency, and generalization capability of the proposed approach.
\end{enumerate}

The remainder of this paper is organized as follows: Section~\ref{Related-works} reviews related works in ED. Section~\ref{Methodology} details the proposed methodology. Section~\ref{Experiments} presents experimental results and analysis. Section~\ref{Discussion} provides discussions. Section~\ref{Conclusion} concludes the paper and discusses future directions.

\section{Related Works}
\label{Related-works}

This section reviews key developments in ED datasets, models, and loss functions, emphasizing how their evolution has shaped current research and motivated this study.

\subsection{Datasets}

Early research on ED was often intertwined with related problems such as contour and boundary detection. Consequently, early studies commonly relied on datasets originally developed for other vision tasks rather than those designed specifically for ED. Among these, the Berkeley Segmentation Dataset (BSDS300) and its extended version BSDS500 \cite{MF2001A}, have been most widely used, containing 300 and 500 RGB images, respectively, each annotated with multiple human-labeled contours. Similarly, the Multi-cue Boundary Dataset (MDBD) \cite{MK2016A} provides 100 high-resolution images with multi-cue annotations, while the NYU Depth Dataset (NYUD) and its extended version NYUDv2 \cite{SH2012Indoor} include 1,449 annotated indoor images. Broader datasets such as PASCAL VOC \cite{EV2010The}, Microsoft COCO \cite{LM2014Microsoft}, SceneParse150 \cite{ZZ2017Scene}, and Cityscapes \cite{CO2016The} have also been adopted due to their visual diversity. However, as these datasets were not explicitly designed for edge delineation, their annotations often lack the precision and consistency required for high-quality ED, thereby constraining task-specific progress.

With ED increasingly recognized as an independent perceptual task, several dedicated datasets have been developed. The BIPED dataset, later refined into BIPED2 \cite{SR2020Dense}, consists of 250 carefully annotated edge images. The BRIND dataset \cite{PH2021RINDNet} extends BSDS500 by categorizing edges into four semantic types, while the UDED dataset \cite{SL2023Tiny} provides 29 images with high-quality edge labels, serving as a compact but reliable benchmark for modern ED algorithms.

A persistent challenge across all ED datasets lies in the noise and inconsistency of human annotations. Ground-truth labels often contain missing or spurious edges, while discrepancies frequently occur among different annotators or even across repeated annotations by the same person. Such inconsistencies complicate both training and evaluation, potentially propagating perceptual errors into downstream applications such as segmentation and recognition. Recent studies have addressed these issues through improved labeling protocols and the design of models robust to noisy supervision \cite{FG2023Practical, WD2024One, S2025Enhancing}. In this paper, we adopt the augmentation method in \cite{S2025Enhancing}, which adds training labels into training data to enhance robustness.

\subsection{Models}

The development of ED models can be broadly divided into heuristic methods, statistical learning schemes, and deep learning approaches.

Early heuristic methods relied on manually defined criteria such as image intensity gradients. Classical detectors like Sobel \cite{K1983On} and Canny \cite{C1986A} remain foundational for their simplicity and efficiency, though their limited adaptability hinders performance across diverse scenes.

Statistical learning methods treated ED as a supervised classification problem using features such as Chernoff information \cite{KY2003Statistical}, local histograms \cite{AM2011Contour}, texture cues \cite{MF2004Learning}, and sketch tokens \cite{LZ2013Sketch}. These features were then combined with classifiers including structured forests \cite{DZ2015Fast}, nearest-neighbor models \cite{GL2014N}, or logistic regression \cite{R2008Multi}. Although these methods improved edge delineation, their heavy reliance on handcrafted features limited scalability and generalization.

The rise of deep learning, particularly convolutional neural networks (CNNs), revolutionized ED through automatic hierarchical feature extraction. Holistically-Nested Edge Detection (HED) \cite{XT2015Holistically} introduced deep supervision and multi-scale fusion, enabling effective edge capture across feature hierarchies. The Richer Convolutional Features (RCF) model \cite{LC2017Richer} further enriched this strategy by exploiting more granular intra-layer features. Later methods such as Bi-Directional Cascade Networks (BDCN) \cite{HZ2022BDCN}, Pixel Difference Networks (PiDiNet) \cite{SL2021Pixel}, and Dense Extreme Inception Networks (Dexi) \cite{SS2023Dense} refined both architectural and supervisory designs to enhance localization precision. More recently, transformer-based frameworks \cite{PH2022Edter, JG2024EdgeNAT} and diffusion-based generative approaches \cite{ZH2024Generative, YX2024Diffusionedge} have shown promise in leveraging global context for perceptual edge reasoning.

Despite these advances, most deep models emphasize feature extraction but underexploit the potential of feature selection. The Extractor–Selector (E-S) framework and its extension, the EES framework \cite{S2025Boost}, addresses this issue by incorporating a selector module that performs pixel-wise feature refinement atop conventional extractors, thereby improving spatial precision. Unlike conventional models that merely fuse deep features (extracted by networks with tens of millions of parameters) using a few shallow convolutional layers, these paradigms employ an auxiliary deep selector network to provide pixel-wise feature refinements. However, despite these architecturally sophisticated designs, such models still struggle to produce perceptually coherent edges that align with human visual judgment—an aspect crucial for reliable perceptual reasoning in intelligent vision systems.

\subsection{Loss Functions}

Beyond architectural innovations, the design of loss functions plays a decisive role in shaping the perceptual behavior of ED models. A central challenge arises from the extreme class imbalance between edge and non-edge pixels. The standard Binary Cross-Entropy (BCE) loss is often inadequate, as it treats both classes equally, leading to biased optimization. To mitigate this, the Weighted Binary Cross-Entropy (WBCE) loss assigns greater importance to edge pixels, thereby improving recall and overall edge accuracy.

However, WBCE-based models typically depend on post-processing methods such as Non-Maximum Suppression (NMS) to enhance visual sharpness. These additional steps disrupt end-to-end optimization and complicate integration into higher-level vision pipelines. To overcome such limitations, several alternative losses have been proposed. The Dice loss \cite{DS2018Learning} balances edge preservation and suppression, while the Tracing loss \cite{HX2022Unmixing} emphasizes edge continuity and texture discrimination by considering spatial proximity to ground truth. Ranking-based losses such as AP-loss \cite{CL2019Towards} and Rank-loss \cite{CK2024RankED} further refine the distinction between edges and textures. Although these formulations improve perceptual quality, they often require complex hyperparameter tuning or sacrifice numerical precision. For example, Tracing loss requires dataset-specific and layer-wise hyperparameter tuning when applied to multi-layer output architectures such as HED and Dexi, while Rank loss incurs nearly 40 GB of memory consumption when training with a single batch of $320 \times 320$ resolution images and can exhibit instability across certain datasets and models, thereby limiting its generalization capability.

As perceptually oriented loss functions tend to introduce trade-offs such as unstable convergence or limited generalization across datasets and architectures, designing a loss function that simultaneously enhances quantitative accuracy and perceptual fidelity—while remaining lightweight, stable, and model-agnostic—remains an open and valuable challenge. For intelligent vision systems, such perceptually consistent optimization is particularly critical, as it directly determines the reliability of the structural cues that support higher-level reasoning and decision-making.

\section{Methodology}
\label{Methodology}

In practical applications, intelligent systems are expected to behave in a human-like manner. Unlike numerical optimization, improving perceptual quality is inherently more challenging, as it lacks an explicit quantitative definition and often depends on heuristic evaluation. Therefore, developing methods that enhance perceptual quality without compromising quantitative accuracy is of both theoretical and practical significance.

In this section, we introduce the Symmetrization Weighted Binary Cross-Entropy (SWBCE) loss, designed to align learning objectives with human perceptual and decision-theoretic characteristics, thereby improving both numerical accuracy and perceptual fidelity in edge detection. We first analyze the limitations of the standard WBCE, then discuss the perceptual motivation inspired by human annotation behavior, and finally derive the proposed formulation.

\subsection{Limitations of WBCE}

The WBCE loss is widely used in ED to address the extreme class imbalance between edge and non-edge pixels. By assigning greater weights to positive (edge) samples, WBCE enhances recall by emphasizing rare edge pixels. However, it often produces blurred and noisy edge maps because pixels near true edges exhibit similar features to edge ones and are therefore prone to misclassification. Yet, WBCE assigns low weights to these perceptually important negative samples, reducing their impact during optimization. As a result, these pixels contribute little to the overall loss, leading to weak suppression of textural regions and ultimately degraded perceptual sharpness.

Therefore, WBCE achieves high recall but often fails to generate visually sharp and perceptually consistent edges, leading to over-prediction and degraded visual quality. When applied to intelligent vision systems, this degradation weakens the reliability of structural features used for high-level reasoning or decision-making.

\subsection{Motivation: Asymmetry in Edge Annotation}

In ED, the high performance hinges on two objectives: accurately detecting true edge pixels (high recall) and avoiding false predicted edges (high precision). False edge predictions are the main source of both texture pollution and edge blurring. However, WBCE mainly focuses on class imbalance and overlooks the perceptual asymmetry underlying human edge recognition, thus failing to effectively suppress false predictions.

Human annotators tend to be cautious when deciding edge pixels. A pixel is typically marked as an edge \textbf{only when there is strong visual evidence}, whereas non-edge labels are assigned by default\footnote{For example, when manually labeling edges in an image, one typically begins by identifying regions that could plausibly contain edges. Then, edges are marked at the pixel level with care and are often reviewed for accuracy, ensuring a high level of confidence before confirming them as true edges. In contrast, regions not marked as edges are usually left unlabelled by default, without explicit verification that each pixel is definitively non-edge.}. This behavior reflects an implicit yet essential principle: edge decisions require explicit evidence, whereas non-edge regions are accepted by default.

This natural asymmetry suggests that models should adopt a more conservative prediction strategy, mimicking human annotation behavior to retain only well-supported edges and achieve perceptual consistency similar to human judgment. However, existing loss functions such as WBCE fail to encode this cautious mechanism. As a result, they often misclassify textural pixels near edges as edges, because these pixels are difficult to suppress yet assigned low weights in the loss. Although perceptually significant, such misclassified pixels contribute little to optimization, ultimately degrading the visual sharpness and consistency of predicted edges.

This gap motivates a perception-aware loss function that explicitly models the asymmetry between edge and non-edge decision-making. Such a loss should:

\begin{itemize}
\item Reflect the cautious nature of edge prediction by penalizing false positives—pixels more heavily.
\item Retain the class imbalance sensitivity to ensure that rare edge pixels remain properly emphasized for high recall.
\end{itemize}

  \subsection{The Symmetrization WBCE (SWBCE) Loss}
  
  To address this issue, we propose the Symmetrization Weighted Binary Cross-Entropy (SWBCE) loss, which augments WBCE with a complementary prediction-based weighting mechanism. SWBCE assumes that an edge should be predicted only when sufficient evidence exists, similar to human decision-making, and penalizes overconfident false positives. It does so by introducing a symmetrical counterpart to the standard WBCE, one that evaluates the loss from the perspective of predictions rather than only the ground-truth.

  Intuitively, SWBCE encourages high recall by emphasizing rare ground-truth edges as in WBCE, while also enforcing high precision by penalizing confidently predicted edges that do not correspond to ground-truth. Since incorrect edge predictions are penalized more severely, models become more cautious and predict edges only when confidence is sufficiently high. This mechanism aligns with the discriminative nature of human perception, thereby improving the perceptual sharpness and structural clarity of predicted edges. Moreover, because texture pollution in ED primarily stems from false edge predictions, the proposed formulation helps generate cleaner and more stable edge maps.

 Formally, SWBCE is defined as:

\begin{equation}
	L_{SWBCE}(\hat{Y},Y)=\frac{L_{Label}(\hat{Y},Y)+b\times L_{Pred}(\hat{Y},Y)}{1+b}
\end{equation}

\noindent where, $L_{Label}$ is the standard WBCE loss function, emphasizing recall of edge pixels,  
 $L_{Pred}$ is the novelly designed prediction-weighted loss, emphasizing precision by penalizing false edges, and $b$ is a hyperparameter balancing $L_{Label}$ and $L_{Pred}$, typically set to 1.

Specifically,
 
\begin{equation}
\begin{aligned}
	L_{Label}(\hat{Y},Y)=&-\sum_{y_{i}\in Y}\beta_{i}[y_{i}log(\hat{y_{i}})+(1-y_{i})log(1-\hat{y_{i}})]\\
    =&-\sum_{y_{i}\in Y^{+}}\beta_{i}log(\hat{y_{i}})-\sum_{y_{i}\in Y^{-}}\beta_{i}log(1-\hat{y_{i}})\\
    =&-\alpha\sum_{y_{i}\in Y^{+}}log(\hat{y_{i}})-\lambda(1-\alpha)\sum_{y_{i}\in Y^{-}}log(1-\hat{y_{i}})
\end{aligned}
\end{equation}

\noindent where, $\hat{y_{i}}$ is the pixel in the prediction $\hat{Y}$ corresponding to the pixel $y_{i}$ in the groundtruth $Y$, $Y^{+}$ is the set of edge pixels in $Y$ (positive samples), $Y^{-}$ is the set of non-edge pixels in $Y$ (negative samples). Here, for $y_{i}\in Y^{+}$, $\beta_{i}=\alpha = \frac{|Y^{-}|}{|Y|}$, while for $y_{i}\in Y^{-}$, $\beta_{i}=\lambda(1-\alpha)$ are their weights. Finally, $\lambda = 1.1$ as suggested in previous works.

\begin{equation}
	L_{Pred}(\hat{Y},Y)=-\sum_{\hat{y_{i}}\in \hat{Y}}\gamma_{i}[y_{i}log(\hat{y_{i}})+(1-y_{i})log(1-\hat{y_{i}})]
\end{equation}

\noindent where, $\gamma_{i}=\hat{y_{i}}\frac{\hat{I}_{N}}{|\hat{Y}|}+(1-\hat{y_{i}})\epsilon\frac{\hat{I}_{P}}{|\hat{Y}|}$ is the weight of the pixel $\hat{y_{i}}$, $\hat{I}_{P}=\sum_{\hat{y_{i}}\in\hat{Y}}\hat{y_{i}}$ is the influence of predicted edges,  $\hat{I}_{N}=|\hat{Y}|-\hat{I}_{P}$ is the influence of predicted non-edges, $\epsilon = 1.1$ is a balancing hyperparameter as $\lambda$ in $L_{Label}$.

Unlike WBCE, which weights samples solely based on their true class labels, SWBCE additionally incorporates prediction-based weighting to achieve symmetric supervision. The former is crucial for high recall, while the latter contributes to high precision. This unified formulation enhances both perceptual sharpness and semantic reliability, rendering the predicted edges more suitable as structured cues for high-level reasoning in vision systems.

The rationale behind $L_{Label}$ (standard WBCE) is to mitigate class imbalance by assigning higher weights to ground-truth edge pixels ($y_i \in Y^+$), thereby ensuring high recall rate for these rare features. In contrast, $L_{Pred}$ introduces a symmetrical supervision mechanism by determining weights based on the model’s own predictive confidence. Specifically, the weight $\gamma_i$ is formulated as a linear interpolation between the normalized count of predicted non-edges $\frac{\hat{I}_{N}}{|\hat{Y}|}$ and predicted edges $\epsilon\frac{\hat{I}_{P}}{|\hat{Y}|}$, governed by the prediction $\hat{y}_i$. Consequently, as a pixel's predicted "edgeness" increases ($\hat{y}_i \to 1$), its weight $\gamma_i$ approaches $\frac{\hat{I}_{N}}{|\hat{Y}|}$; conversely, for high-confidence non-edge predictions $\hat{y}_i \to 0$, $\gamma_i$ approaches $\epsilon\frac{\hat{I}_{P}}{|\hat{Y}|}$. Given that $\hat{I}_{N} \gg \hat{I}_{P}$ in natural images, this formulation explicitly models the asymmetric confirmation bias inherent to human vision. 

From the perspective of human decision-making under uncertainty, identifying a structural edge requires strong evidence, whereas the absence of such evidence defaults to a background (non-edge) assumption. By making the penalty weight $\gamma_i$ proportional to the predicted probability, $L_{Pred}$ effectively acts as a risk-sensitive regulator. It imposes a high cost on overconfident false positives—instances where the model asserts an edge without sufficient evidence. This mechanism encourages the model to become more conservative and evidence-driven, mirroring the cautious nature of human visual confirmation. Mathematically, the symmetry between label-based recall ($L_{Label}$) and prediction-based precision ($L_{Pred}$) establishes a dual-constraint optimization objective, bridging statistical class balance and perceptual reliability in ED.

\section{Experiments}
\label{Experiments}

In this section, we conduct extensive experiments to evaluate the effectiveness of the proposed SWBCE loss, benchmarking it against recent alternatives across multiple architectures and datasets. We first introduce an additional evaluation criterion to assess model predictions beyond conventional metrics, followed by a description of the experimental benchmarks. We then present the main experimental results along with ablation studies on parameter stability.

\subsection{Edge Ratio}

We define a new evaluation criterion, termed \emph{Edge Ratio} (ER), for assessing edge predictions. ER serves as a continuous analogue of the precision rate, enabling quantitative evaluation without binarization.

\begin{equation}
ER := \frac{\text{Sum of predicted pixel values within the edge region}}{\text{Sum of all predicted pixel values}}
\end{equation}

Here, the edge region is defined as the set of non-zero pixels obtained by convolving the ground truth with a $3 \times 3$ kernel whose elements are all ones. Specifically, a pixel is considered to lie within the edge region if it falls inside a $3 \times 3$ neighborhood centered at any ground-truth edge pixel. Accordingly, ER can be interpreted as follows: given a total predicted mass of $a$ distributed across the image, and the correctly distributed mass is $b$ concentrated on edge regions, ER is defined as the ratio $\frac{b}{a}$. A higher ER therefore indicates higher precision, corresponding to cleaner and less noisy edge predictions.

When predictions are binary maps, ER reduces to the conventional pixel-level correct rate. However, since ED often serves as a low-level cue extraction module and may be jointly trained with high-level vision tasks, binarization is frequently undesirable since it is non-continuous and thus incompatible with gradient-based optimization pipelines. In this context, ER provides a more general criterion for evaluating ED results.

\subsection{Benchmarks and Notations}

\subsubsection{Baseline}

The proposed SWBCE loss is evaluated against three representative loss functions: the long-adopted WBCE loss, the Tracing loss \cite{HX2022Unmixing} presented in 2022, and the Rank loss \cite{CK2024RankED} presented in 2024.

These loss functions are evaluated across four ED architectures: HED \cite{XT2015Holistically}, BDCN \cite{HZ2022BDCN}, Dexi \cite{SS2023Dense}, and EdgeNat\cite{JG2024EdgeNAT}. The main experiments on HED, BDCN, and Dexi are within the EES framework \cite{S2025Boost} presented in 2025. Evaluations without the EES framework can be found in the Appendix.

Experiments are conducted on four public datasets, including BIPED2, UDED, and BRIND, which are specifically for ED tasks, as well as NYUD2, which is ordinary for segmentation tasks. This configuration covers both dedicated edge benchmarks and tasks where edges serve as intermediate perceptual cues, providing evidence that SWBCE improves both ED itself and its application to high-level tasks.

\subsubsection{Evaluation Protocol}

We adopt standard evaluation metrics following \cite{MF2004Learning}, including the Optimal Dataset Scale (ODS), Optimal Image Scale (OIS), and Average Precision (AP), to quantitatively assess ED performance. Following the recommendation of \cite{S2025Enhancing}, all evaluations are conducted under a 1-pixel error tolerance, which imposes a significantly more stringent accuracy requirement than the commonly used 4- to 11-pixel error-tolerance range\footnote{The 1-pixel tolerance corresponds to an error distance threshold of 0.003 for BRIND and UDED, 0.002 for NYUD2, and 0.001 for BIPED2 set in the standard algorithm presented in \cite{MF2004Learning}. For comparison, the default 0.0075 tolerance would allow up to 4.3, 11.1, and 6 pixels error tolerance for BRIND, BIPED2, and NYUD2, respectively, leading to less precise evaluation.}. Furthermore, structural similarity index measure (SSIM), and root mean squared error (RMSE) are also evaluated. All evaluations are conducted without post-processing techniques such as NMS, since these non-differentiable operations cannot be integrated into high-level models for joint training, thus reducing the usability of edges as structured cues for higher-level reasoning. Results under traditional error tolerance with NMS can be found in the Appendix.

\subsubsection{Training Details}

For the EES framework, we adopt a two-stage training strategy, comprising extractor pre-training and selector training, following \cite{S2025Boost}. The union fine-tuning stage is omitted to save resources, as both quantitative scores and perceptual qualities are sufficiently achieved through the first two stages. Other procedures are similar to those in \cite{S2025Boost}:

\textbf{Data processing:} Each training image is recursively split into halves until both its height and width are below 640 pixels. An exception is made for NYUD2, which is not split, accounting for its data number, to prevent it from containing many more images than others, and save computational cost. Training data is augmented via 4 rotations and horizontal flipping, and noiseless data is also included. During training, $320 \times 320$ patches are randomly cropped and refreshed every 5 epochs to increase diversity. Predictions are made on patches of $320\times 320$ resolution and merged post hoc.

\textbf{Training hyperparameters:} Learning rate is set to $10^{-4}$ with weight decay set to $10^{-8}$. The batch size is 8 for WBCE, tracing, and SWBCE loss, while 1 for rank loss\footnote{Rank loss incurs high GPU memory usage (up to 40 GB for a single batch), making larger batch sizes impractical.}. For UDED, 25 epochs are run for rank loss, and 200 for other losses, while for other datasets, 7 epochs are run for rank loss, and 50 for others\footnote{This is to approximately equalize the total number of optimization steps across loss functions given their differing batch sizes.}.

\textbf{Dataset partition:} BIPED2 (250 images) is split to 200/50, UDED (27 used images) is split to 20/7, BRIND (500 images) is split to 400/100, and NYUD2 (1400 used images) is split to 1100/300 training/testing images, respectively.

All models are trained and evaluated under consistent settings to isolate the effect of loss functions on performance.

\subsection{Experiment Results}

\begin{figure}[htbp]
        \centering
        \includegraphics[width=\textwidth]{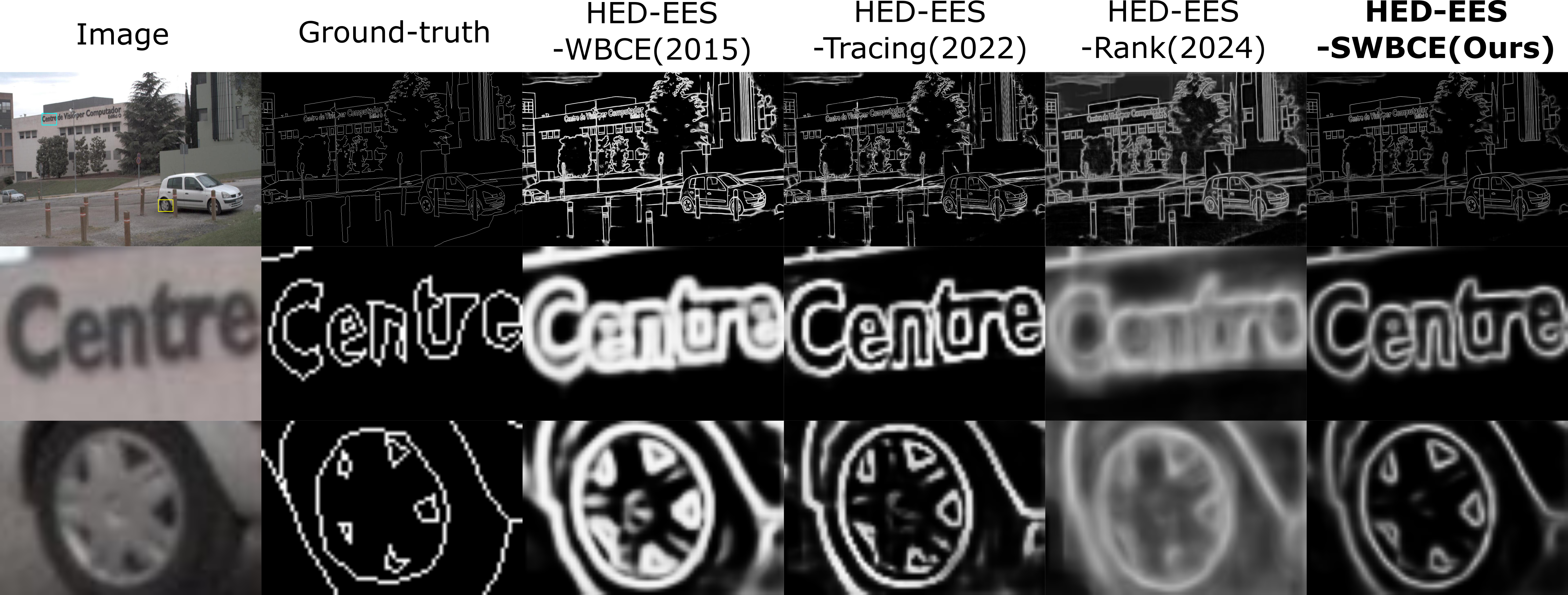}
        \caption{\ \textbf{Qualitative comparisons on HED-EES} for the four loss functions. Although tracing loss also provides clear results, it preserves more unwanted artifacts, such as in the middle of the tire (bottom), while the proposed method yields cleaner results, filtering such artifacts.}
        \label{fig:HEDEES}
    \end{figure}

\begin{figure}[htbp]
        \centering
        \includegraphics[width=\textwidth]{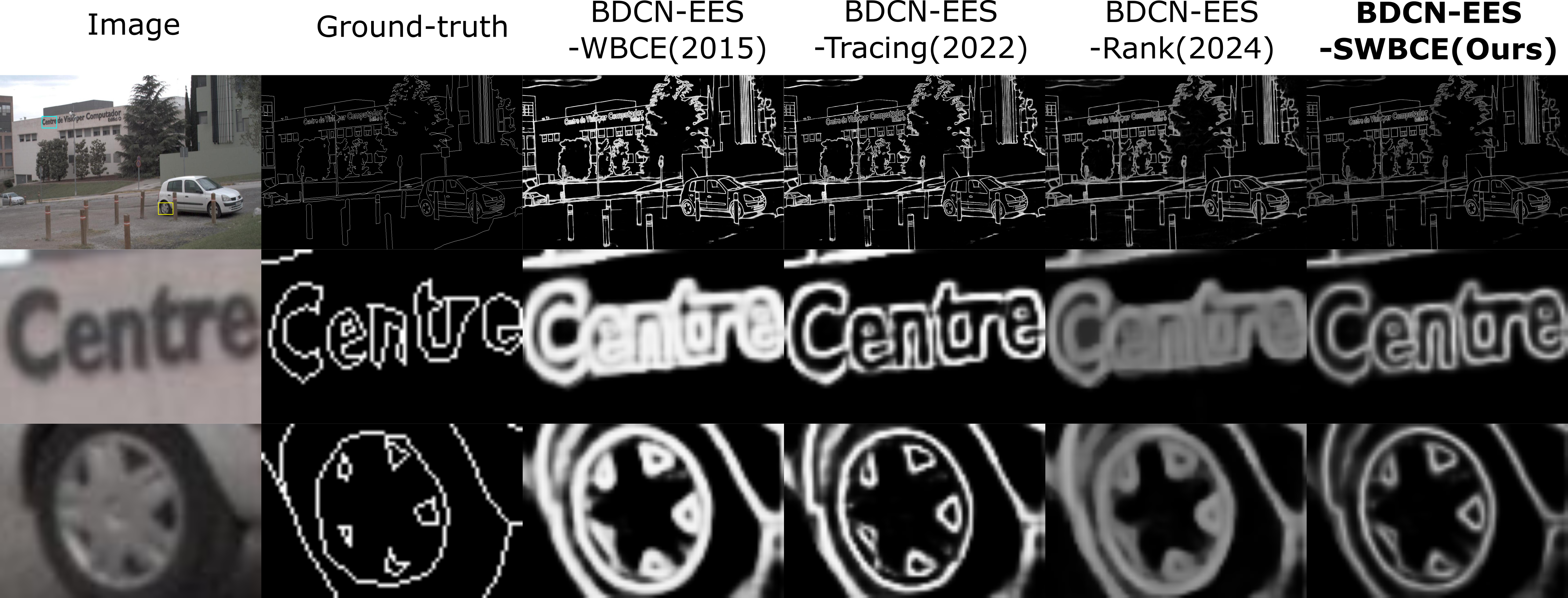}
        \caption{\ \textbf{Qualitative comparisons on BDCN-EES} for the four loss functions. While the tracing loss avoids the pollution in the middle of the tire (bottom), SWBCE still yields more perceptual results, which is thinner and smoother, consistent with the performance on HED-EES.}
        \label{fig:BDCNEES}
    \end{figure}

\begin{figure}[htbp]
        \centering
        \includegraphics[width=\textwidth]{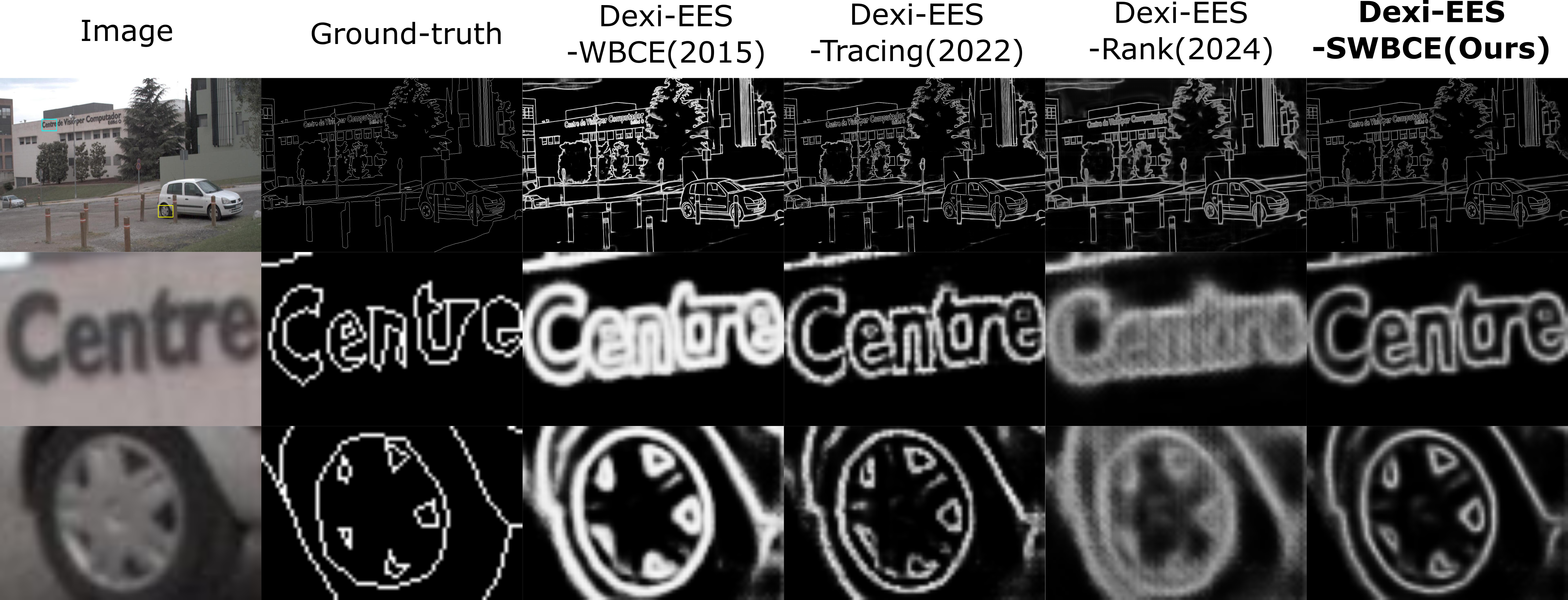}
        \caption{\ \textbf{Qualitative comparisons on Dexi-EES} for the four loss functions. Previous loss functions introduce artifacts, such as in the middle of the tire (bottom), and glitches in the words (middle), while SWBCE yields more perceptual and smooth results.}
        \label{fig:DexiEES}
    \end{figure}

\begin{figure}[htbp]
        \centering
        \includegraphics[width=\textwidth]{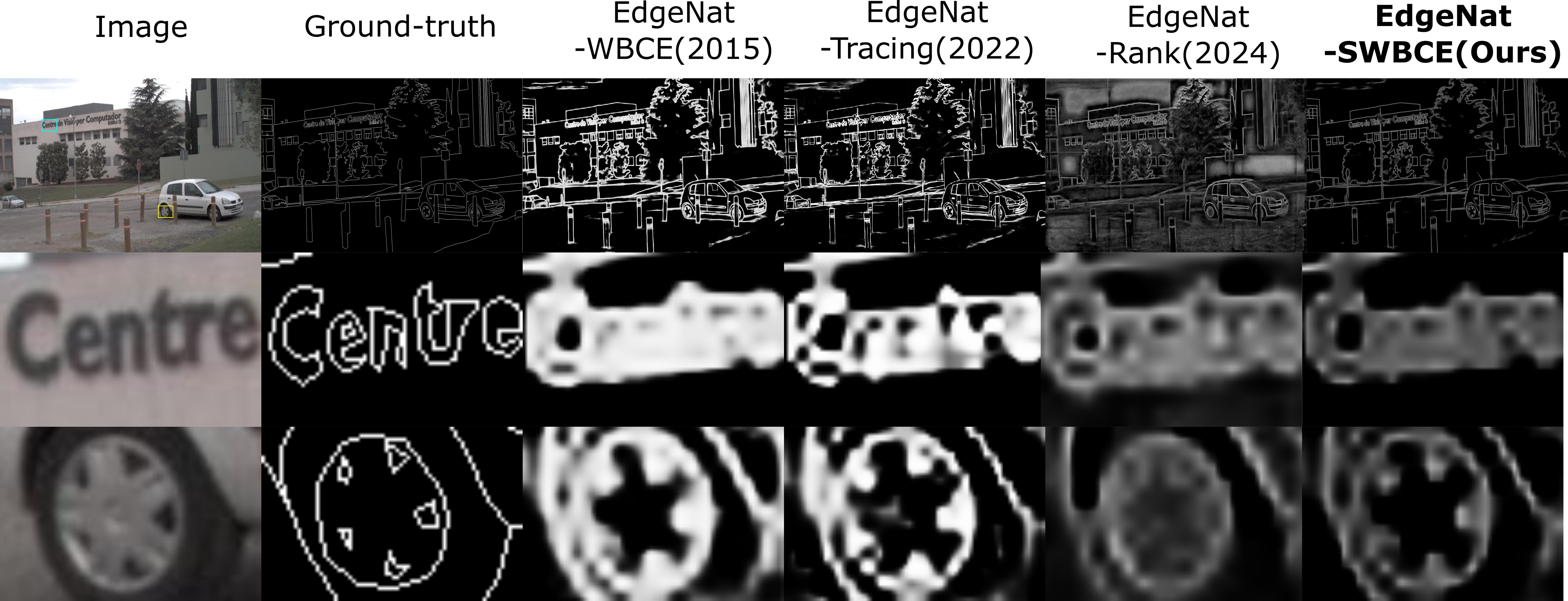}
        \caption{\ \textbf{Qualitative comparisons on EdgeNat} for the four loss functions. All loss function seems not to perform well, indicating that this model may not be suitable for this dataset (BIPED2). However, SWBCE results still look better.}
        \label{fig:EdgeNat}
    \end{figure}

Quantitative results are summarized in Tables \ref{HED-EES-BB} to \ref{EdgeNat-UN}, and representative qualitative predictions are visualized in Fig.\ref{fig:HEDEES} to \ref{fig:EdgeNat}. The best scores are marked in \textcolor{blue}{blue} and the second best are marked in \textcolor{red}{red}. The proposed SWBCE loss consistently achieves comparable or superior quantitative performance across datasets and baselines, and produces notably more perceptual, cleaner, and crisper visual results compared to previous loss functions.

\begin{table}[htbp]
\renewcommand\arraystretch{0.8}
\centering
\caption{\textbf{Results of the HED-EES on BIPED2 and BRIND, with 1-pixel error tolerance without NMS.} On the two datasets, training with SWBCE consistently obtains the best quantitative scores, in all five criteria. Particularly, it improves the SSIM on BIPED2 and BRIND by about 8.0\% and 15.5\%, respectively, compared to the second-best loss.}
\label{HED-EES-BB}
\small
\begin{tabular}{|@{\hspace{0.5mm}}c@{\hspace{0.5mm}}|@{\hspace{1mm}}c@{\hspace{1.5mm}}c@{\hspace{1.5mm}}c@{\hspace{1.5mm}}c@{\hspace{1.5mm}}c@{\hspace{0.5mm}}c@{\hspace{0.5mm}}|@{\hspace{1mm}}c@{\hspace{1.5mm}}c@{\hspace{1.5mm}}c@{\hspace{1.5mm}}c@{\hspace{1.5mm}}c@{\hspace{0.5mm}}c@{\hspace{0.5mm}}|}
 \hline
 &\multicolumn{6}{c}{BIPED2}&\multicolumn{6}{c|}{BRIND}
 \\
\hline
    & ODS   & OIS   & AP  & SSIM & ER & RMSE & ODS   & OIS   & AP & SSIM & ER &RMSE \\
\hline
WBCE  & \textcolor{red}{0.666} & \textcolor{red}{0.671} & \textcolor{red}{0.493} & 0.574& \textcolor{red}{0.466}& 0.301&\textcolor{red}{0.679} & \textcolor{red}{0.687} & 0.492 & \textcolor{red}{0.420}& \textcolor{red}{0.418}&0.312
\\
Tracing & 0.655 & 0.661 & 0.484  &\textcolor{red}{0.576} & 0.461& \textcolor{red}{0.230}& 0.659 & 0.674 & \textcolor{red}{0.568}  & 0.403& 0.400&0.241
\\
Rank & 0.560 & 0.571 & 0.270  & 0.056& 0.241&0.269& 0.656 & 0.671 & 0.530  & 0.237& 0.333&\textcolor{red}{0.210}\\
SWBCE & \textcolor{blue}{0.667} & \textcolor{blue}{0.674} & \textcolor{blue}{0.543}  & \textcolor{blue}{0.622}& \textcolor{blue}{0.523}&\textcolor{blue}{0.169}& \textcolor{blue}{0.681} & \textcolor{blue}{0.691} & \textcolor{blue}{0.594}  & \textcolor{blue}{0.485}&\textcolor{blue}{0.468}& \textcolor{blue}{0.175}\\
\hline
\end{tabular}
\end{table}

\begin{table}[htbp]
\renewcommand\arraystretch{0.8}
\centering
\caption{\textbf{Results of the HED-EES on UDED and NYU2, with 1-pixel error tolerance without NMS.} Training with SWBCE can obtain the best scores in most cases. Particularly, the ODS on UDED and NYU2 improves by about 4.9\% and 4.6\%, respectively, compared to the second-best ones.}
\label{HED-EES-UN}
\small
\begin{tabular}{|@{\hspace{0.5mm}}c@{\hspace{0.5mm}}|@{\hspace{1mm}}c@{\hspace{1.5mm}}c@{\hspace{1.5mm}}c@{\hspace{1.5mm}}c@{\hspace{1.5mm}}c@{\hspace{0.5mm}}c@{\hspace{0.5mm}}|@{\hspace{1mm}}c@{\hspace{1.5mm}}c@{\hspace{1.5mm}}c@{\hspace{1.5mm}}c@{\hspace{1.5mm}}c@{\hspace{0.5mm}}c@{\hspace{0.5mm}}|}
 \hline
 &\multicolumn{6}{c}{UDED}&\multicolumn{6}{c|}{NYU2}
 \\
\hline
    & ODS   & OIS   & AP  & SSIM & ER & RMSE & ODS   & OIS   & AP  & SSIM & ER & RMSE \\
\hline
WBCE    & \textcolor{red}{0.738} & \textcolor{red}{0.765} & 0.635 &\textcolor{blue}{0.422}&\textcolor{blue}{0.548}&0.308 & \textcolor{red}{0.409} & \textcolor{red}{0.418} & \textcolor{red}{0.154}&\textcolor{red}{0.307}&\textcolor{red}{0.254}&0.405
\\
Tracing  & \textcolor{red}{0.738} & 0.757 & \textcolor{blue}{0.713} &0.351&0.435& 0.280& 0.406 & 0.413 & 0.148 &\textcolor{red}{0.307}&\textcolor{red}{0.254}&0.350
\\
Rank  & 0.613 & 0.646 & 0.447 &0.228&0.372&\textcolor{red}{0.239} & 0.400 & 0.409 & 0.137 &0.058&0.204&\textcolor{red}{0.322}\\
SWBCE  & \textcolor{blue}{0.774} & \textcolor{blue}{0.787} & \textcolor{red}{0.661} &\textcolor{red}{0.384}&\textcolor{red}{0.530}&\textcolor{blue}{0.223}& \textcolor{blue}{0.428} & \textcolor{blue}{0.436} & \textcolor{blue}{0.200} &\textcolor{blue}{0.352}&\textcolor{blue}{0.287}&\textcolor{blue}{0.269}\\
\hline
\end{tabular}
\end{table}

\begin{table}[htbp]
\renewcommand\arraystretch{0.8}
\centering
\caption{\textbf{Results of the BDCN-EES on BIPED2 and BRIND with 1-pixel error tolerance without NMS.} Training with SWBCE can obtain the best score in most cases. Particularly, on BIPED2 and BRIND, the SSIM is improved by about 6.9\% and 15.6\%, respectively, compared to the second-best ones.}
\label{BDCN-EES-BB}
\small
\begin{tabular}{|@{\hspace{0.5mm}}c@{\hspace{0.5mm}}|@{\hspace{1mm}}c@{\hspace{1.5mm}}c@{\hspace{1.5mm}}c@{\hspace{1.5mm}}c@{\hspace{1.5mm}}c@{\hspace{0.5mm}}c@{\hspace{0.5mm}}|@{\hspace{1mm}}c@{\hspace{1.5mm}}c@{\hspace{1.5mm}}c@{\hspace{1.5mm}}c@{\hspace{1.5mm}}c@{\hspace{0.5mm}}c@{\hspace{0.5mm}}|}
 \hline
 &\multicolumn{6}{c}{BIPED2}&\multicolumn{6}{c|}{BRIND}
 \\
\hline
    & ODS   & OIS   & AP  & SSIM & ER & RMSE & ODS   & OIS   & AP & SSIM & ER &RMSE \\
\hline
WBCE 
& \textcolor{red}{0.652} & 0.655 & 0.473&0.625&0.504&0.294
& \textcolor{blue}{0.678} & \textcolor{red}{0.686} & 0.521&0.499&0.458&0.299
\\
Tracing 
& \textcolor{red}{0.652} & \textcolor{red}{0.659} & \textcolor{red}{0.505} &\textcolor{red}{0.653}&\textcolor{red}{0.536}&0.255
& \textcolor{red}{0.674} & \textcolor{blue}{0.688} & \textcolor{blue}{0.580}&\textcolor{red}{0.555}&\textcolor{red}{0.498}&0.252
\\
Rank 
& 0.622 & 0.630 & 0.416&0.511&0.475&\textcolor{red}{0.203}
& 0.672 & 0.683 & 0.518&0.384&0.415&\textcolor{red}{0.226}
\\
SWBCE 
& \textcolor{blue}{0.662} & \textcolor{blue}{0.666} & \textcolor{blue}{0.515} &\textcolor{blue}{0.668}&\textcolor{blue}{0.559}&\textcolor{blue}{0.171}
& 0.673 & 0.684 & \textcolor{red}{0.566}&\textcolor{blue}{0.577}&\textcolor{blue}{0.508}&\textcolor{blue}{0.171}
\\
\hline
\end{tabular}
\end{table}

\begin{table}[htbp]
\renewcommand\arraystretch{0.8}
\centering
\caption{\textbf{Results of the BDCN-EES on UDED and NYU2 with 1-pixel error tolerance without NMS.} In most cases, training with SWBCE obtains the best scores. Particularly, on UDED and NYU2, the SSIM improves by about 5.2\% and 10.6\%, respectively, compared to the second-best one. }
\label{BDCN-EES-UN}
\small
\begin{tabular}{|@{\hspace{0.5mm}}c@{\hspace{0.5mm}}|@{\hspace{1mm}}c@{\hspace{1.5mm}}c@{\hspace{1.5mm}}c@{\hspace{1.5mm}}c@{\hspace{1.5mm}}c@{\hspace{0.5mm}}c@{\hspace{0.5mm}}|@{\hspace{1mm}}c@{\hspace{1.5mm}}c@{\hspace{1.5mm}}c@{\hspace{1.5mm}}c@{\hspace{1.5mm}}c@{\hspace{0.5mm}}c@{\hspace{0.5mm}}|}
 \hline
 &\multicolumn{6}{c}{UDED}&\multicolumn{6}{c|}{NYU2}
 \\
\hline
    & ODS   & OIS   & AP & SSIM & ER &RMSE& ODS   & OIS   & AP & SSIM & ER &RMSE
    \\
    \hline
WBCE 
& 0.750 & \textcolor{red}{0.777} & \textcolor{red}{0.628}  &\textcolor{red}{0.537}&\textcolor{red}{0.574}&0.304
& \textcolor{red}{0.408} & 0.417 & 0.142&0.311&0.273&0.399
\\
Tracing 
& 0.735 & 0.758 & 0.598&0.543&0.584&0.301
& \textcolor{blue}{0.414} & \textcolor{blue}{0.424} & 0.167 &\textcolor{red}{0.323}&0.286&0.368
\\
Rank 
& \textcolor{blue}{0.755} & 0.770 & 0.558&0.128&0.397&\textcolor{red}{0.274}
& 0.406 & 0.414 & \textcolor{red}{0.171} &0.306&\textcolor{red}{0.287}&\textcolor{red}{0.313}
\\
SWBCE 
& \textcolor{red}{0.752} & \textcolor{blue}{0.782} & \textcolor{blue}{0.670}&\textcolor{blue}{0.565}&\textcolor{blue}{0.601}&\textcolor{blue}{0.217}
& \textcolor{red}{0.408} & \textcolor{red}{0.419} & \textcolor{blue}{0.181}&\textcolor{blue}{0.344}&\textcolor{blue}{0.293}&\textcolor{blue}{0.307}\\
\hline
\end{tabular}
\end{table}

\begin{table}[htbp]
\renewcommand\arraystretch{0.8}
\centering
\caption{\textbf{Results of the Dexi-EES on BIPED2 and BRIND, with 1-pixel error tolerance without NMS.} Training with SWBCE can obtain the best scores in most cases. Particularly, on BIPED2 and BRIND, the SSIM is improved by about 10.1\% and 22.4\%, respectively, compared to the second-best ones.}
\label{Dexi-EES-BB}
\small
\begin{tabular}{|@{\hspace{0.5mm}}c@{\hspace{0.5mm}}|@{\hspace{1mm}}c@{\hspace{1.5mm}}c@{\hspace{1.5mm}}c@{\hspace{1.5mm}}c@{\hspace{1.5mm}}c@{\hspace{0.5mm}}c@{\hspace{0.5mm}}|@{\hspace{1mm}}c@{\hspace{1.5mm}}c@{\hspace{1.5mm}}c@{\hspace{1.5mm}}c@{\hspace{1.5mm}}c@{\hspace{0.5mm}}c@{\hspace{0.5mm}}|}
 \hline
 &\multicolumn{6}{c}{BIPED2}&\multicolumn{6}{c|}{BRIND}
 \\
\hline
    & ODS   & OIS   & AP  & SSIM & ER & RMSE & ODS   & OIS   & AP & SSIM & ER &RMSE \\
\hline
WBCE 
 & \textcolor{red}{0.672} & \textcolor{red}{0.677} & 0.499&0.603&0.504&0.284
& \textcolor{blue}{0.683} & \textcolor{blue}{0.691} & 0.490&\textcolor{red}{0.415}&\textcolor{red}{0.430}&0.300
\\
Tracing 
& 0.667 & 0.671 & \textcolor{red}{0.537}&\textcolor{red}{0.641}&\textcolor{red}{0.540}&\textcolor{red}{0.220}
& 0.663 & 0.673 & \textcolor{blue}{0.583}&0.409&0.427&\textcolor{red}{0.232}
\\
Rank 
& 0.636 & 0.640 & 0.399&0.247&0.361&0.226
& 0.663 & 0.675 & 0.449&0.089&0.288&0.244
\\
SWBCE 
& \textcolor{blue}{0.677} & \textcolor{blue}{0.681} & \textcolor{blue}{0.540}&\textcolor{blue}{0.664}&\textcolor{blue}{0.564}&\textcolor{blue}{0.170}
& \textcolor{red}{0.682} & \textcolor{red}{0.689} & \textcolor{red}{0.566}&\textcolor{blue}{0.508}&\textcolor{blue}{0.482}&\textcolor{blue}{0.174}
\\
\hline
\end{tabular}
\end{table}

\begin{table}[htbp]
\renewcommand\arraystretch{0.8}
\centering
\caption{\textbf{Results of the Dexi-EES on UDED and NYU2 with 1-pixel error tolerance without NMS.} Training with SWBCE can obtain the best scores in most cases. Particularly, on UDED and NYU2, the SSIM is improved by about 12.2\% and 14.1\%, respectively, compared to the second-best ones.}
\label{Dexi-EES-UN}
\small
\begin{tabular}{|@{\hspace{0.5mm}}c@{\hspace{0.5mm}}|@{\hspace{1mm}}c@{\hspace{1.5mm}}c@{\hspace{1.5mm}}c@{\hspace{1.5mm}}c@{\hspace{1.5mm}}c@{\hspace{0.5mm}}c@{\hspace{0.5mm}}|@{\hspace{1mm}}c@{\hspace{1.5mm}}c@{\hspace{1.5mm}}c@{\hspace{1.5mm}}c@{\hspace{1.5mm}}c@{\hspace{0.5mm}}c@{\hspace{0.5mm}}|}
 \hline
 &\multicolumn{6}{c}{UDED}&\multicolumn{6}{c|}{NYU2}
 \\
\hline
    & ODS   & OIS   & AP & SSIM & ER &RMSE& ODS   & OIS   & AP & SSIM & ER &RMSE
    \\
    \hline
WBCE 
& \textcolor{blue}{0.784} & \textcolor{red}{0.796} & \textcolor{red}{0.683} &\textcolor{red}{0.449}&0.526&0.314
& 0.402 & 0.413 & 0.158&0.318&0.262&0.390
\\
Tracing 
& 0.761 & 0.782 & 0.744&0.429&\textcolor{red}{0.531}&\textcolor{red}{0.251}
& \textcolor{red}{0.417} & \textcolor{red}{0.427} & \textcolor{red}{0.162} &\textcolor{red}{0.335}&\textcolor{red}{0.276}&0.356
\\
Rank 
& 0.764 & 0.775 & 0.526&0.095&0.386&0.264
& 0.410 & 0.420 & 0.155 &0.184&0.240&\textcolor{red}{0.342}
\\
SWBCE 
& \textcolor{red}{0.780} & \textcolor{blue}{0.801} & \textcolor{blue}{0.720}&\textcolor{blue}{0.504}&\textcolor{blue}{0.590}&\textcolor{blue}{0.217}
& \textcolor{blue}{0.428} & \textcolor{blue}{0.438} & \textcolor{blue}{0.201}&\textcolor{blue}{0.363}&\textcolor{blue}{0.291}&\textcolor{blue}{0.283}\\
\hline
\end{tabular}
\end{table}

\begin{table}[htbp]
\renewcommand\arraystretch{0.8}
\centering
\caption{\textbf{Results of the EdgeNat on BIPED2 and BRIND, with 1-pixel error tolerance without NMS.} Training with SWBCE can obtain the best scores in most cases. Particularly, on BIPED2 and BRIND, the SSIM is improved by about 12.1\% and 6.4\%, respectively, compared to the second-best ones.}
\label{EdgeNat-BB}
\small
\begin{tabular}{|@{\hspace{0.5mm}}c@{\hspace{0.5mm}}|@{\hspace{1mm}}c@{\hspace{1.5mm}}c@{\hspace{1.5mm}}c@{\hspace{1.5mm}}c@{\hspace{1.5mm}}c@{\hspace{0.5mm}}c@{\hspace{0.5mm}}|@{\hspace{1mm}}c@{\hspace{1.5mm}}c@{\hspace{1.5mm}}c@{\hspace{1.5mm}}c@{\hspace{1.5mm}}c@{\hspace{0.5mm}}c@{\hspace{0.5mm}}|}
 \hline
 &\multicolumn{6}{c}{BIPED2}&\multicolumn{6}{c|}{BRIND}
 \\
\hline
    & ODS   & OIS   & AP & SSIM & ER & RMSE & ODS   & OIS   & AP  & SSIM & ER & RMSE
    \\
\hline
WBCE & \textcolor{red}{0.593}& \textcolor{red}{0.595} & \textcolor{red}{0.376} &\textcolor{red}{0.592}&0.462&0.313 & \textcolor{blue}{0.612} & \textcolor{red}{0.621} & 0.411  & \textcolor{red}{0.488} & 0.435 & 0.314
\\
Tracing & 0.586 & 0.590 & 0.317  &0.557&\textcolor{red}{0.469}&\textcolor{red}{0.286}& 0.604 & 0.614 & \textcolor{red}{0.426}  & 0.468 & \textcolor{red}{0.445} & \textcolor{red}{0.276}
\\
Rank & 0.268 & 0.298 & 0.120  &0.036&0.143&0.321& 0.509 & 0.527 & 0.369   & 0.066 & 0.197 & 0.293
\\
SWBCE & \textcolor{blue}{0.596} & \textcolor{blue}{0.599} & \textcolor{blue}{0.401}  &\textcolor{blue}{0.664}&\textcolor{blue}{0.537}&\textcolor{blue}{0.172}& \textcolor{red}{0.607} & \textcolor{blue}{0.623} & \textcolor{blue}{0.433}  & \textcolor{blue}{0.519} & \textcolor{blue}{0.469} & \textcolor{blue}{0.182}
\\
\hline
\end{tabular}
\end{table}

\begin{table}[htbp]
\renewcommand\arraystretch{0.8}
\centering
\caption{\textbf{Results of the EdgeNat on UDED and NYU2, with 1-pixel error tolerance without NMS.} Training with SWBCE can obtain the best scores in most cases. Particularly, on UDED and NYU2, the AP is improved by about 8.4\% and 24.4\%, respectively, compared to the second-best ones.}
\label{EdgeNat-UN}
\small
\begin{tabular}{|@{\hspace{0.5mm}}c@{\hspace{0.5mm}}|@{\hspace{1mm}}c@{\hspace{1.5mm}}c@{\hspace{1.5mm}}c@{\hspace{1.5mm}}c@{\hspace{1.5mm}}c@{\hspace{0.5mm}}c@{\hspace{0.5mm}}|@{\hspace{1mm}}c@{\hspace{1.5mm}}c@{\hspace{1.5mm}}c@{\hspace{1.5mm}}c@{\hspace{1.5mm}}c@{\hspace{0.5mm}}c@{\hspace{0.5mm}}|}
 \hline
 &\multicolumn{6}{c}{UDED}&\multicolumn{6}{c|}{NYU2}
 \\
\hline
    & ODS   & OIS   & AP & SSIM   & ER   & RMSE & ODS   & OIS   & AP & SSIM   & ER   & RMSE
    \\
\hline
WBCE   & 0.664 & 0.668 & \textcolor{red}{0.474} &\textcolor{red}{0.468}&\textcolor{red}{0.521}&0.343 & 0.407 & \textcolor{red}{0.416} & 0.167&0.304&0.250&0.408
\\
Tracing  & \textcolor{blue}{0.678} & \textcolor{blue}{0.683} & 0.448 &\textcolor{blue}{0.497} & \textcolor{blue}{0.530} & 0.320 & \textcolor{red}{0.409} & \textcolor{red}{0.416} & 0.170 &0.238&0.256&0.362
\\
Rank  & 0.617 & 0.635 & 0.452 & 0.390& 0.405& \textcolor{red}{0.276}& 0.365 & 0.377 & \textcolor{red}{0.176} &\textcolor{blue}{0.394}&\textcolor{red}{0.292}&\textcolor{red}{0.247}
\\
SWBCE  & \textcolor{red}{0.670} & \textcolor{red}{0.676} & \textcolor{blue}{0.490} &0.428&0.484&\textcolor{blue}{0.230} & \textcolor{blue}{0.422} & \textcolor{blue}{0.431} & \textcolor{blue}{0.219} &\textcolor{red}{0.393} &\textcolor{blue}{0.308}&\textcolor{blue}{0.242}
\\
\hline
\end{tabular}
\end{table}

\subsection{Ablation}

\begin{figure}[htbp]
        \centering
        \begin{subfigure}{0.32\columnwidth}
             \includegraphics[width=\linewidth]{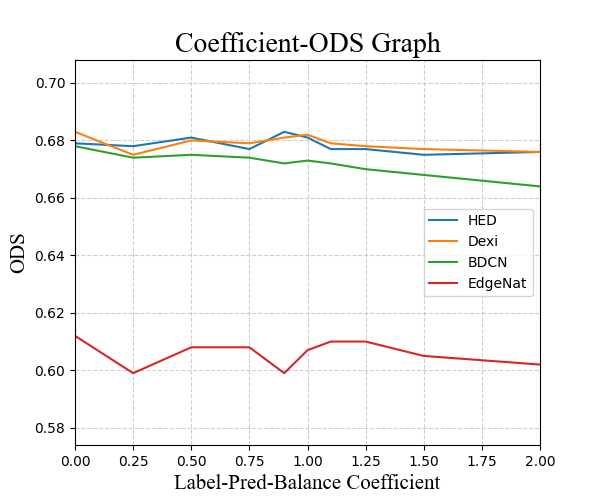}
             \caption{ODS vs.\ $b$}
        \end{subfigure}
        \hfill
        \begin{subfigure}{0.32\columnwidth}
             \includegraphics[width=\linewidth]{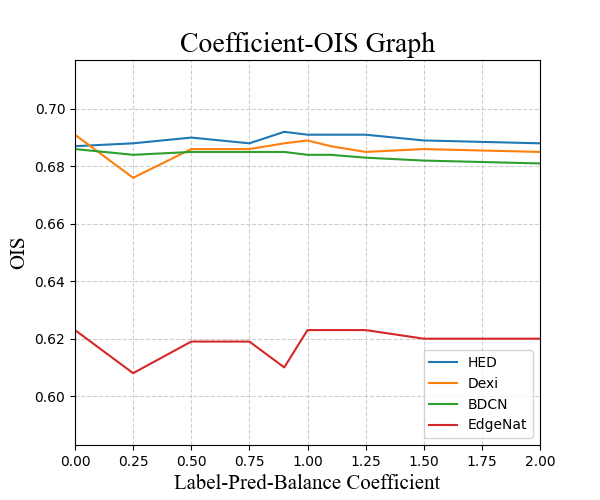}
             \caption{OIS vs.\ $b$}
        \end{subfigure}
        \hfill
        \begin{subfigure}{0.32\columnwidth}
             \includegraphics[width=\linewidth]{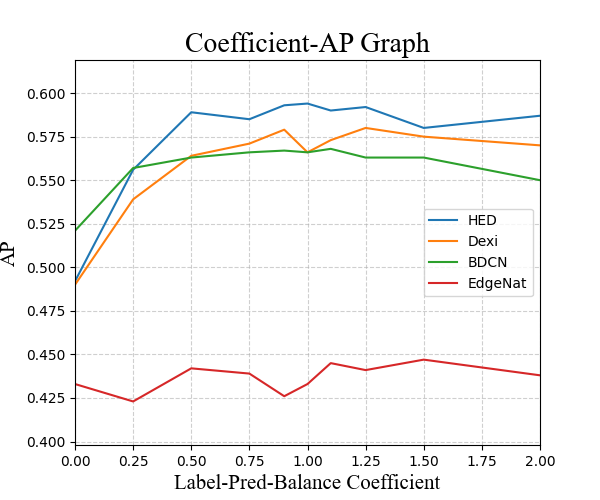}
             \caption{AP vs.\ $b$}
        \end{subfigure}
        \caption{\ Stability of the hyperparameter $b$ balancing between $L_{Label}$ and $L_{Pred}$ on BRIND}
        \label{fig:BRIND-b}
    \end{figure}

We evaluate the impact of the balancing hyperparameter between the two components of the loss, $L_{Label}$ and $L_{Pred}$. As shown in Fig.\ref{fig:BRIND-b} to \ref{fig:UDED-b}, the performance is relatively stable across a wide range of settings, suggesting robustness to this parameter, which ensures that SWBCE can be reliably deployed without model-specific tuning. Table \ref{BRIND-b} and \ref{UDED-b} provide full details of the experiments, adjusting values of the balancing hyperparameter $b$ between the two loss components.

Moreover, it should be noted that using $L_{Pred}$ alone (i.e., $b = +\infty$) leads to very poor performance. This behavior is expected, as the weighting scheme in $L_{Pred}$ is derived solely from model predictions and is independent of the ground-truth, which informs the true edge distribution. During early training stages, when the model initialization yields low confidence for true edge pixels, the corresponding weights remain small, making these pixels difficult to optimize despite being severely mispredicted. This observation indicates that $L_{Pred}$ should be regarded as a complementary component within a composite loss formulation, rather than being used in isolation.

    \begin{figure}[htbp]
        \centering
        \begin{subfigure}{0.32\columnwidth}
             \includegraphics[width=\linewidth]{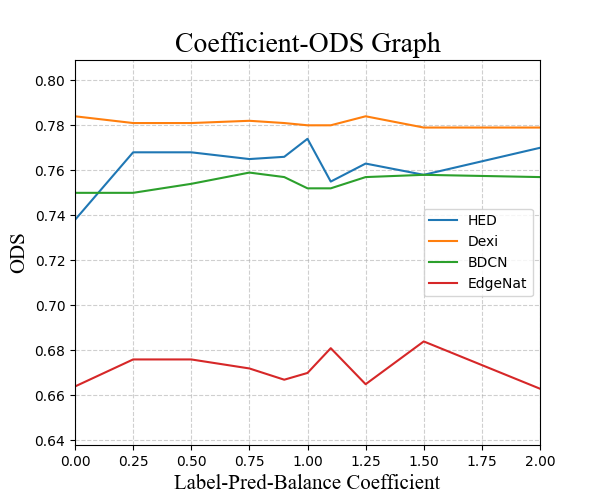}
             \caption{ODS vs.\ $b$}
        \end{subfigure}
        \hfill
        \begin{subfigure}{0.32\columnwidth}
             \includegraphics[width=\linewidth]{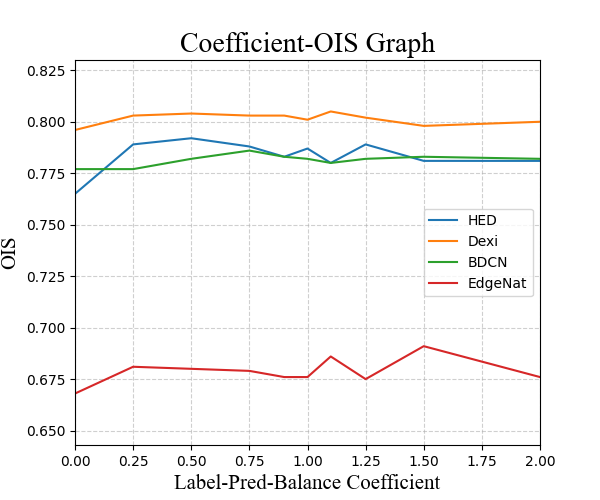}
             \caption{OIS vs.\ $b$}
        \end{subfigure}
        \hfill
        \begin{subfigure}{0.32\columnwidth}
             \includegraphics[width=\linewidth]{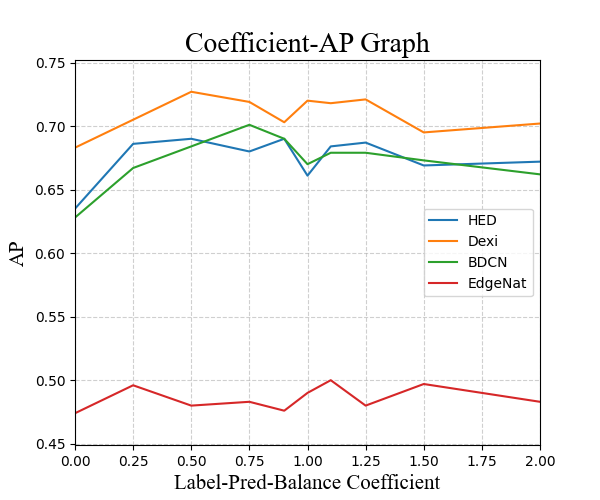}
             \caption{AP vs.\ $b$}
        \end{subfigure}
        \caption{\ Stability of the hyperparameter $b$ balancing between $L_{Label}$ and $L_{Pred}$ on UDED}
        \label{fig:UDED-b}
    \end{figure}

    \begin{figure}[htbp]
        \centering
        \begin{subfigure}{0.32\columnwidth}
             \includegraphics[width=\linewidth]{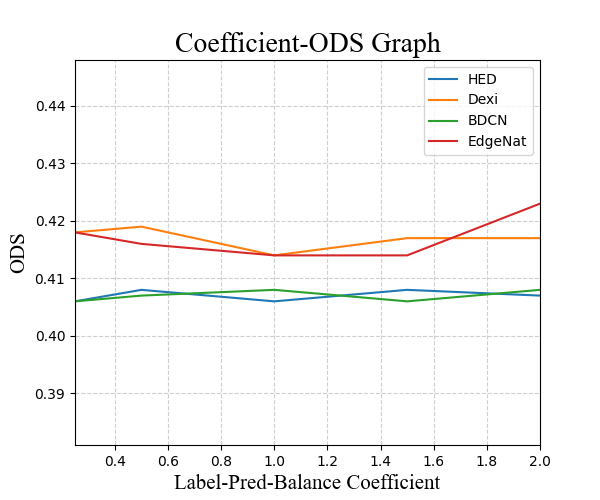}
             \caption{ODS vs.\ $b$}
        \end{subfigure}
        \hfill
        \begin{subfigure}{0.32\columnwidth}
             \includegraphics[width=\linewidth]{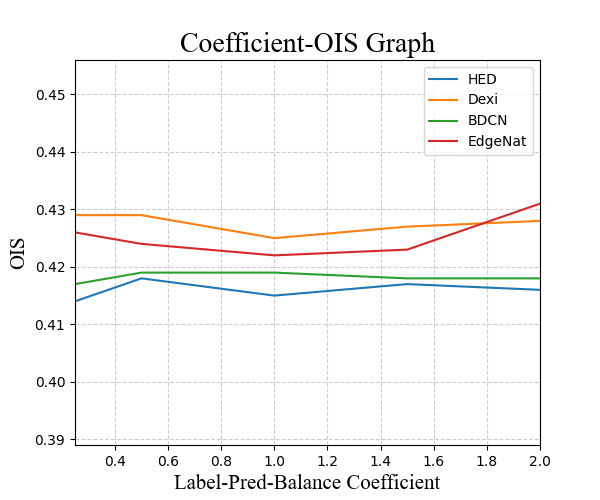}
             \caption{OIS vs.\ $b$}
        \end{subfigure}
        \hfill
        \begin{subfigure}{0.32\columnwidth}
             \includegraphics[width=\linewidth]{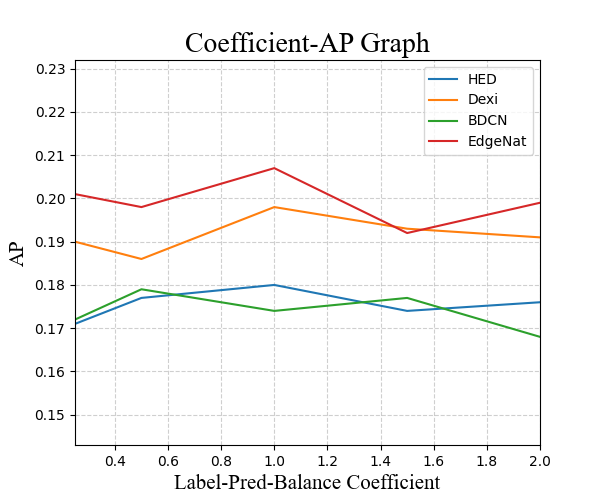}
             \caption{AP vs.\ $b$}
        \end{subfigure}
        \caption{\ Stability of the hyperparameter $b$ balancing between $L_{Label}$ and $L_{Pred}$ on NYU2}
        \label{fig:UDED-b}
    \end{figure}

\begin{table}[htbp]
	\renewcommand\arraystretch{0.8}
	\centering
	\caption{\textbf{Hyperparameter Study:} Effect of different balancing parameter $b$ of SWBCE, on BRIND, 1-pixel error tolerance, and without NMS. $b=0$ represents using WBCE only, whose results are provided in the main experiments. $b=+\infty$ represents using $L_{Pred}$ only, which is unsatisfactory since the importance of the pixels is highly dependent on the initialized prediction, which may result in lots of unwanted abortion.}
	\label{BRIND-b}
	\begin{tabular}{|@{\hspace{1mm}}c@{\hspace{1mm}}|@{\hspace{1mm}}c@{\hspace{1.5mm}}c@{\hspace{1.5mm}}c@{\hspace{1mm}}|@{\hspace{1mm}}c@{\hspace{1.5mm}}c@{\hspace{1.5mm}}c@{\hspace{1mm}}|@{\hspace{1mm}}c@{\hspace{1.5mm}}c@{\hspace{1.5mm}}c@{\hspace{1mm}}|@{\hspace{1mm}}c@{\hspace{1.5mm}}c@{\hspace{1.5mm}}c@{\hspace{1mm}}|}
\hline 
 &\multicolumn{3}{c}{HED-EES}&\multicolumn{3}{c}{BDCN-EES}&\multicolumn{3}{c}{Dexi-EES}&\multicolumn{3}{c|}{EdgeNat}\\
\hline
$b$  & ODS   & OIS   & AP& ODS   & OIS   & AP& ODS   & OIS   & AP& ODS   & OIS   & AP 
\\
\hline
0.25 & 0.678 & 0.688 & 0.556 & 0.674 & 0.684 & 0.557 & 0.665 & 0.676 & 0.539 & 0.599 & 0.608 & 0.423
\\
0.5 & 0.681 & 0.690 & 0.589 & 0.675 & 0.685 & 0.563 & 0.680 & 0.686 & 0.564  & 0.608 & 0.619 & 0.442 
\\
0.75 & 0.677 & 0.688 & 0.585 & 0.674 & 0.685 & 0.566 & 0.679 & 0.686 & 0.571  & 0.608 & 0.619 & 0.439
\\
0.9 & 0.683 & 0.692 & 0.593 & 0.672 & 0.685 & 0.567 & 0.681 & 0.688 & 0.579  & 0.599 & 0.610 & 0.426 
\\
1 & 0.681 & 0.691 & 0.594 & 0.673 & 0.684 & 0.566 & 0.682 & 0.689 & 0.566  & 0.607 & 0.623 & 0.433
\\
1.1 & 0.677 & 0.691 & 0.590 & 0.672 & 0.684 & 0.568 & 0.679 & 0.687 & 0.573  & 0.610 & 0.623 & 0.445 
\\
1.25 & 0.677 & 0.691 & 0.592 & 0.670 & 0.683 & 0.563 & 0.678 & 0.685 & 0.580  & 0.610 & 0.623 & 0.441 \\
1.5 & 0.675 & 0.689 & 0.580 & 0.668 & 0.682 & 0.563 & 0.677 & 0.686 & 0.575  & 0.605 & 0.620 & 0.447 
\\
2 & 0.676 & 0.688 & 0.587 & 0.664 & 0.681 & 0.550 & 0.676 & 0.685 & 0.570  & 0.602 & 0.620 & 0.438
\\
$+\infty$ & 0.179 & 0.194 & 0.020 & 0.270 & 0.297 & 0.115 & 0.181 & 0.199 & 0.021  & 0.149 & 0.157 & 0.011
\\
\hline
\end{tabular}  
\end{table}

\begin{table}[htbp]
	\renewcommand\arraystretch{0.8}
	\centering
	\caption{\textbf{Hyperparameter Study:} Effect of different balancing parameter $b$ of SWBCE, on UDED, 1-pixel error tolerance, and without NMS.}
	\label{UDED-b}
	\begin{tabular}{|@{\hspace{1mm}}c@{\hspace{1mm}}|@{\hspace{1mm}}c@{\hspace{1.5mm}}c@{\hspace{1.5mm}}c@{\hspace{1mm}}|@{\hspace{1mm}}c@{\hspace{1.5mm}}c@{\hspace{1.5mm}}c@{\hspace{1mm}}|@{\hspace{1mm}}c@{\hspace{1.5mm}}c@{\hspace{1.5mm}}c@{\hspace{1mm}}|@{\hspace{1mm}}c@{\hspace{1.5mm}}c@{\hspace{1.5mm}}c@{\hspace{1mm}}|}
\hline 
 &\multicolumn{3}{c}{HED-EES}&\multicolumn{3}{c}{BDCN-EES}&\multicolumn{3}{c}{Dexi-EES}&\multicolumn{3}{c|}{EdgeNat}\\
\hline
$b$  & ODS   & OIS   & AP & ODS   & OIS   & AP & ODS   & OIS   & AP & ODS   & OIS   & AP \\
\hline
0.25 & 0.768 & 0.789 & 0.686 & 0.750 & 0.777 & 0.667 & 0.781 & 0.803 & 0.705 & 0.676 & 0.681 & 0.496
\\
0.5 & 0.768 & 0.792 & 0.690 & 0.754 & 0.782 & 0.684 & 0.781 & 0.804 & 0.727 & 0.676 & 0.680 & 0.480
\\
0.75 & 0.765 & 0.788 & 0.680 & 0.759 & 0.786 & 0.701 & 0.782 & 0.803 & 0.719 & 0.672 & 0.679 & 0.483
\\
0.9 & 0.766 & 0.783 & 0.690 & 0.757 & 0.783 & 0.690 & 0.781 & 0.803 & 0.703 & 0.667 & 0.676 & 0.476
\\
1 & 0.774 & 0.787 & 0.661 & 0.752 & 0.782 & 0.670 & 0.780 & 0.801 & 0.720  & 0.670 & 0.676 & 0.490 
\\
1.1 & 0.755 & 0.780 & 0.684 & 0.752 & 0.780 & 0.670 & 0.780 & 0.805 & 0.718 & 0.681 & 0.686 & 0.500
\\
1.25 & 0.763 & 0.789 & 0.687 & 0.757 & 0.782 & 0.679 & 0.784 & 0.802 & 0.721 & 0.665 & 0.675 & 0.480
\\
1.5 & 0.758 & 0.781 & 0.669 & 0.758 & 0.783 & 0.673 & 0.779 & 0.798 & 0.695 & 0.684 & 0.691 & 0.497
\\
2 & 0.770 & 0.781 & 0.672 & 0.757 & 0.782 & 0.662 & 0.779 & 0.800 & 0.702 & 0.663 & 0.676 & 0.483
\\
$+\infty$ & 0.326 & 0.362 & 0.128 & 0.573 & 0.601 & 0.392 & 0.392 & 0.462 & 0.185 & 0.236 & 0.255 & 0.036
\\
\hline
\end{tabular}  
\end{table}

\begin{table}[htbp]
	\renewcommand\arraystretch{0.8}
	\centering
	\caption{\textbf{Hyperparameter Study:} Effect of different balancing parameter $b$ of SWBCE, on NYU2, 1-pixel error tolerance, and without NMS. Epoch 25 is used.}
	\label{NYU2C-b}
	\begin{tabular}{|@{\hspace{1mm}}c@{\hspace{1mm}}|@{\hspace{1mm}}c@{\hspace{1.5mm}}c@{\hspace{1.5mm}}c@{\hspace{1mm}}|@{\hspace{1mm}}c@{\hspace{1.5mm}}c@{\hspace{1.5mm}}c@{\hspace{1mm}}|@{\hspace{1mm}}c@{\hspace{1.5mm}}c@{\hspace{1.5mm}}c@{\hspace{1mm}}|@{\hspace{1mm}}c@{\hspace{1.5mm}}c@{\hspace{1.5mm}}c@{\hspace{1mm}}|}
\hline 
 &\multicolumn{3}{c}{HED-EES}&\multicolumn{3}{c}{BDCN-EES}&\multicolumn{3}{c}{Dexi-EES}&\multicolumn{3}{c|}{EdgeNat}\\
\hline
$b$  & ODS   & OIS   & AP& ODS   & OIS   & AP& ODS   & OIS   & AP& ODS   & OIS   & AP 
\\
\hline
0.25 & 0.406 & 0.414 & 0.171 & 0.406 & 0.417 & 0.172 & 0.418 & 0.429 & 0.190 & 0.418 & 0.426 & 0.201
\\
0.5 & 0.408 & 0.418 & 0.177 & 0.407 & 0.419 & 0.179 & 0.419 & 0.429 & 0.186  & 0.416 & 0.424 & 0.198 
\\
1 & 0.406 & 0.415 & 0.180 & 0.408 & 0.419 & 0.174 & 0.414 & 0.425 & 0.198  & 0.414 & 0.422 & 0.207
\\
1.5 & 0.408 & 0.417 & 0.174 & 0.406 & 0.418 & 0.177 & 0.417 & 0.427 & 0.193  & 0.414 & 0.423 & 0.192 
\\
2 & 0.407 & 0.416 & 0.176 & 0.408 & 0.418 & 0.168 & 0.417 & 0.428 & 0.191  & 0.423 & 0.431 & 0.199
\\
\hline
\end{tabular}  
\end{table}

\section{Discussion}
\label{Discussion}

We provide auxiliary discussions for the proposed SWBCE loss. 

First, as shown in the experimental results, SWBCE does not consistently achieve the best quantitative performance across all datasets. Although performance inconsistency across datasets is common for neural networks and loss functions, this variability remains an issue that warrants further investigation.

Second, as illustrated in Fig.~\ref{fig:EdgeNat}, SWBCE does not always yield perceptually superior results. Its effectiveness partially depends on the baseline performance of WBCE, which is expected since SWBCE inherits half of its supervisory signal from the standard WBCE formulation. When a model performs poorly under WBCE, it may fall outside the effective improvement range of SWBCE. Consequently, additional experiments are required to more clearly characterize the applicable scope and limitations of the proposed loss. Notably, such dependency on baseline behavior is a common challenge in neural network optimization and loss design, and remains a long-standing open problem.

Third, SWBCE introduces additional computational overhead during training. According to training records on HED-EED using the BRIND dataset with 6,400 images (after data augmentation), a batch size of 8, and an input resolution of $320 \times 320$, SWBCE requires approximately 391 seconds per epoch, compared to 294 seconds per epoch for WBCE—representing an increase of about 35\% on an RTX A6000 GPU. This overhead mainly arises from the auxiliary computation of the prediction-based loss term $L_{Pred}$. Although the additional cost affects only the training stage and not inference, it nonetheless introduces a trade-off between computational efficiency and perceptual performance. For comparison, Tracing loss requires approximately 404 seconds per epoch with a batch size of 8, while Rank loss requires about 864 seconds per epoch with a batch size of 1. In terms of memory consumption, WBCE, Tracing, and SWBCE require approximately 11,710 MB during training with a batch size of 8, whereas Rank loss requires 26,358 MB with a batch size of 1, indicating that SWBCE will not cause a significant memory burden compared to WBCE.

\section{Conclusion}
\label{Conclusion}

This paper presented the Symmetrization Weighted Binary Cross-Entropy (SWBCE) loss, a perception-inspired formulation for edge detection (ED) that explicitly models the asymmetry inherent in human visual decision-making. By integrating prediction-guided weighting into the conventional WBCE framework, SWBCE establishes a symmetric and perceptually consistent optimization mechanism that enhances edge recall while suppressing false positives, achieving a human-like balance between quantitative accuracy and perceptual fidelity.

Comprehensive experiments across diverse datasets and architectures confirm its robustness and generalization capability. Particularly, with the HED-EES model, the SSIM can be improved by about 15\% on BRIND, and in all experiments, training by SWBCE consistently obtains the best perceptual results. Most importantly, this work demonstrates that explicitly modeling perceptual asymmetry bridges the gap between statistical optimization and human perceptual reasoning, providing a foundation for perceptually grounded loss design, which could potentially benefit a wide range of computational tasks.

\bibliographystyle{unsrt}
\bibliography{EDBitex}

\newpage

\section{Appendix}

This section provides supplementary quantitative analyses under conventional relaxed evaluation protocols and additional ablation studies.

\subsection{Experimental Results on the Traditional Relaxed Criterion}

Tables \ref{HED-EES-75}, \ref{BDCN-EES-75}, \ref{Dexi-EES-75}, and \ref{EdgeNat-75} present evaluation results using the traditional relaxed benchmark, where the error tolerance is set to 0.0075 times the diagonal length, following the evaluation protocol in \cite{MF2004Learning}. This corresponds approximately to a tolerance of 4.3, 11.1, and 6 pixels for BRIND, BIPED2, and NYUD2, respectively. All results include NMS. The notations follow the main text.

\begin{table}[htbp]
\renewcommand\arraystretch{0.8}
\centering
\caption{\textbf{Results of HED-EES on traditional error tolerance with NMS.}}
\label{HED-EES-75}
\small
\begin{tabular}{|@{\hspace{1mm}}c@{\hspace{1mm}}|@{\hspace{1mm}}c@{\hspace{1.5mm}}c@{\hspace{1.5mm}}c@{\hspace{1mm}}|@{\hspace{1mm}}c@{\hspace{1.5mm}}c@{\hspace{1.5mm}}c@{\hspace{1mm}}|@{\hspace{1mm}}c@{\hspace{1.5mm}}c@{\hspace{1.5mm}}c@{\hspace{1mm}}|@{\hspace{1mm}}c@{\hspace{1.5mm}}c@{\hspace{1.5mm}}c@{\hspace{1mm}}|}
 \hline
 &\multicolumn{3}{c}{BIPED2}&\multicolumn{3}{c}{BRIND}&\multicolumn{3}{c}{UDED}&\multicolumn{3}{c|}{NYU2}
 \\
\hline
    & ODS   & OIS   & AP & ODS   & OIS   & AP & ODS   & OIS   & AP & ODS   & OIS   & AP 
    \\
\hline
WBCE 
& \textcolor{red}{0.882} & \textcolor{red}{0.893} & \textcolor{blue}{0.927} 
& \textcolor{red}{0.789} & \textcolor{red}{0.803} & \textcolor{blue}{0.849} 
& \textcolor{red}{0.823} & \textcolor{red}{0.845} & \textcolor{red}{0.856}
& \textcolor{red}{0.710} & \textcolor{red}{0.721} & \textcolor{blue}{0.687}
\\
Tracing 
& 0.875 & 0.884 & 0.913 
& 0.774 & 0.793 & 0.812 
& 0.813 & 0.831 & 0.851
& 0.708 & 0.718 & 0.662 
\\
Rank 
& 0.869 & 0.887 & 0.927 
& 0.780 & 0.798 & \textcolor{red}{0.843} 
& 0.750 & 0.784 & 0.786
& \textcolor{red}{0.710} & 0.720 & 0.619  
\\
SWBCE 
& \textcolor{blue}{0.885} & \textcolor{blue}{0.894} & \textcolor{red}{0.926}
& \textcolor{blue}{0.796} & \textcolor{blue}{0.808} & \textcolor{red}{0.843} 
& \textcolor{blue}{0.852} & \textcolor{blue}{0.862} & \textcolor{blue}{0.890}
& \textcolor{blue}{0.713} & \textcolor{blue}{0.722} & \textcolor{red}{0.672}
\\
\hline
\end{tabular}
\end{table}

\begin{table}[htbp]
\renewcommand\arraystretch{0.8}
\centering
\caption{\textbf{Results of BDCN-EES on traditional error tolerance with NMS.}}
\label{BDCN-EES-75}
\small
\begin{tabular}{|@{\hspace{1mm}}c@{\hspace{1mm}}|@{\hspace{1mm}}c@{\hspace{1.5mm}}c@{\hspace{1.5mm}}c@{\hspace{1mm}}|@{\hspace{1mm}}c@{\hspace{1.5mm}}c@{\hspace{1.5mm}}c@{\hspace{1mm}}|@{\hspace{1mm}}c@{\hspace{1.5mm}}c@{\hspace{1.5mm}}c@{\hspace{1mm}}|@{\hspace{1mm}}c@{\hspace{1.5mm}}c@{\hspace{1.5mm}}c@{\hspace{1mm}}|}
 \hline
 &\multicolumn{3}{c}{BIPED2}&\multicolumn{3}{c}{BRIND}&\multicolumn{3}{c}{UDED}&\multicolumn{3}{c|}{NYU2}
 \\
\hline
    & ODS   & OIS   & AP & ODS   & OIS   & AP & ODS   & OIS   & AP & ODS   & OIS   & AP 
    \\
\hline
WBCE 
  & \textcolor{red}{0.881} & \textcolor{red}{0.890} & \textcolor{blue}{0.921}
& \textcolor{blue}{0.790} & \textcolor{blue}{0.806} & \textcolor{red}{0.831} 
& 0.825 & \textcolor{blue}{0.854} & 0.822
& \textcolor{red}{0.720} & \textcolor{red}{0.732} & \textcolor{red}{0.708}
\\
Tracing 
& 0.870 & 0.881 & 0.898 
& 0.786 & \textcolor{red}{0.802} & 0.801 
& \textcolor{red}{0.828} & \textcolor{red}{0.853} & 0.831
& \textcolor{blue}{0.722} & \textcolor{blue}{0.733} & \textcolor{blue}{0.728} 
\\
Rank 
& 0.878 & 0.888 & \textcolor{blue}{0.921} 
& 0.786 & \textcolor{red}{0.802} & \textcolor{blue}{0.843} 
& \textcolor{blue}{0.832} & \textcolor{blue}{0.854} & \textcolor{blue}{0.888}
& \textcolor{red}{0.720} & 0.731 & 0.681  
\\
SWBCE 
& \textcolor{blue}{0.884} & \textcolor{red}{0.890} & \textcolor{red}{0.920} 
& \textcolor{red}{0.787} & 0.801 & 0.819 
& 0.820 & \textcolor{red}{0.853} & \textcolor{red}{0.853}
& \textcolor{red}{0.720} & 0.731 & \textcolor{red}{0.708}\\
\hline
\end{tabular}
\end{table}

\begin{table}[htbp]
\renewcommand\arraystretch{0.8}
\centering
\caption{\textbf{Results of Dexi-EES on traditional error tolerance with NMS.}}
\label{Dexi-EES-75}
\small
\begin{tabular}{|@{\hspace{1mm}}c@{\hspace{1mm}}|@{\hspace{1mm}}c@{\hspace{1.5mm}}c@{\hspace{1.5mm}}c@{\hspace{1mm}}|@{\hspace{1mm}}c@{\hspace{1.5mm}}c@{\hspace{1.5mm}}c@{\hspace{1mm}}|@{\hspace{1mm}}c@{\hspace{1.5mm}}c@{\hspace{1.5mm}}c@{\hspace{1mm}}|@{\hspace{1mm}}c@{\hspace{1.5mm}}c@{\hspace{1.5mm}}c@{\hspace{1mm}}|}
 \hline
 &\multicolumn{3}{c}{BIPED2}&\multicolumn{3}{c}{BRIND}&\multicolumn{3}{c}{UDED}&\multicolumn{3}{c|}{NYU2}
 \\
\hline
    & ODS   & OIS   & AP & ODS   & OIS   & AP & ODS   & OIS   & AP & ODS   & OIS   & AP 
    \\
\hline
WBCE 
  & \textcolor{red}{0.887} & \textcolor{red}{0.895} & \textcolor{blue}{0.934}
& \textcolor{red}{0.792} & \textcolor{blue}{0.806} & \textcolor{blue}{0.842}
& \textcolor{red}{0.852} & \textcolor{red}{0.869} & 0.889 
& 0.634 & 0.644 & 0.652
\\
Tracing 
& 0.885 & 0.891 & 0.928
& 0.777 & 0.790 & 0.826
& 0.835 & 0.859 & 0.893
& 0.663 & 0.673 & 0.682 
\\
Rank 
& 0.883 & 0.890 & 0.924
& 0.785 & 0.799 & \textcolor{blue}{0.842}
& 0.848 & 0.864 & \textcolor{blue}{0.904}
& \textcolor{red}{0.715} & \textcolor{red}{0.726} & \textcolor{red}{0.731}  
\\
SWBCE 
& \textcolor{blue}{0.888} & \textcolor{blue}{0.896} & \textcolor{red}{0.929}
& \textcolor{blue}{0.794} & \textcolor{red}{0.804} & \textcolor{red}{0.835}
& \textcolor{blue}{0.853} & \textcolor{blue}{0.876} & \textcolor{red}{0.903}
& \textcolor{blue}{0.727} & \textcolor{blue}{0.739} & \textcolor{blue}{0.750}\\
\hline
\end{tabular}
\end{table}

\begin{table}[htbp]
\renewcommand\arraystretch{0.8}
\centering
\caption{\textbf{Results of EdgeNat on traditional error tolerance with NMS.}}
\label{EdgeNat-75}
\small
\begin{tabular}{|@{\hspace{1mm}}c@{\hspace{1mm}}|@{\hspace{1mm}}c@{\hspace{1.5mm}}c@{\hspace{1.5mm}}c@{\hspace{1mm}}|@{\hspace{1mm}}c@{\hspace{1.5mm}}c@{\hspace{1.5mm}}c@{\hspace{1mm}}|@{\hspace{1mm}}c@{\hspace{1.5mm}}c@{\hspace{1.5mm}}c@{\hspace{1mm}}|@{\hspace{1mm}}c@{\hspace{1.5mm}}c@{\hspace{1.5mm}}c@{\hspace{1mm}}|}
 \hline
 &\multicolumn{3}{c}{BIPED2}&\multicolumn{3}{c}{BRIND}&\multicolumn{3}{c}{UDED}&\multicolumn{3}{c|}{NYU2}
 \\
\hline
    & ODS   & OIS   & AP & ODS   & OIS   & AP & ODS   & OIS   & AP & ODS   & OIS   & AP \\
\hline
WBCE & \textcolor{blue}{0.877}& \textcolor{blue}{0.885} & \textcolor{blue}{0.885} & \textcolor{blue}{0.783} & \textcolor{red}{0.801} & \textcolor{blue}{0.794}  & \textcolor{blue}{0.796} & \textcolor{blue}{0.802} & \textcolor{blue}{0.752}  & \textcolor{red}{0.703}  & \textcolor{red}{0.713} & \textcolor{blue}{0.680}
\\
Tracing & \textcolor{red}{0.872} & \textcolor{red}{0.880} & 0.764 & \textcolor{red}{0.779} & \textcolor{blue}{0.796} & 0.767 & \textcolor{red}{0.789} & \textcolor{red}{0.795} & 0.647  & 0.700 & 0.710 & \textcolor{red}{0.678} 
\\
Rank & 0.514 & 0.579 & 0.561 & 0.646 & 0.690 & 0.701 & 0.739 & 0.757 & 0.698  & 0.667 & 0.684 & 0.642 \\
SWBCE & \textcolor{red}{0.872} & \textcolor{red}{0.880} & \textcolor{red}{0.874} & 0.774 & \textcolor{blue}{0.796} & \textcolor{red}{0.774} & 0.782 & 0.788 & \textcolor{red}{0.735}  & \textcolor{blue}{0.711} & \textcolor{blue}{0.719} & 0.671 \\
\hline
\end{tabular}
\end{table}

\subsection{Results on Previous Architectures Without EES Framework}

We also evaluate the proposed losses by applying them directly to prior models without incorporating the EES framework (see Table \ref{HED-BIPED2-BRIND}, \ref{BDCN-BIPED2-BRIND}, and \ref{Dexi-BIPED2-BRIND}). To avoid memory overflow, only the final layer of each model is used for supervision with the rank loss, unlike prior works that leverage multi-level outputs. For the tracing loss and the SWBCE loss, both last-layer and multi-level supervision strategies are examined, while for the WBCE loss, default layer weights are directly employed since they have been fine-tuned in previous works.

In the case of SWBCE with multi-level supervision, we use the same layer weights optimized for the tracing loss, which may not be optimal for SWBCE. This explains why, in some cases, using only the final output yields better performance. We did not fine-tune these hyperparameters for direct application, as the EES-enhanced version shows significantly better performance and is the focus of the main text. Nonetheless, this direct evaluation provides a coarse but informative comparison of loss functions. For a fair comparison, we recommend referring to either the EES implementations or last-layer-only setups. The experimental results also demonstrate the effectiveness of the SWBCE.

The notations follow the main text, and \textit{-Last} denotes supervision using only the final output layer, as opposed to multi-level supervision.

\begin{table}[htbp]
\renewcommand\arraystretch{0.8}
\centering
\caption{\textbf{Results of HED on BIPED2 and BRIND with 1-pixel error tolerance without NMS.}}
\label{HED-BIPED2-BRIND}
\small
\begin{tabular}{|@{}c@{}|@{\hspace{0.5mm}}c@{\hspace{1mm}}c@{\hspace{1mm}}c@{\hspace{1mm}}c@{\hspace{1mm}}c@{\hspace{0.5mm}}c@{}|@{\hspace{0.5mm}}c@{\hspace{1mm}}c@{\hspace{1mm}}c@{\hspace{1mm}}c@{\hspace{1mm}}c@{\hspace{0.5mm}}c@{}|}
 \hline
 &\multicolumn{6}{c}{BIPED2}&\multicolumn{6}{c|}{BRIND}
 \\
\hline
    & ODS   & OIS   & AP &SSIM&ER&RMSE & ODS   & OIS   & AP &SSIM&ER&RMSE
    \\
\hline
WBCE & 0.592 & 0.602 & 0.347&\textcolor{red}{0.578}&0.442&0.319& 0.645 & 0.656 & 0.417  &0.490&0.431&0.314
\\
Tracing & 0.604 & 0.610 & 0.415&0.552&0.473&0.267 & 0.656 & 0.667 & 0.563  &\textcolor{red}{0.518}&\textcolor{blue}{0.509}&0.239
\\
\textbf{SWBCE} & 0.583 & 0.590 & 0.374 &0.547 &0.454&\textcolor{red}{0.198}&0.650 & 0.661 & 0.518  &0.484&0.470&\textcolor{red}{0.179}
\\
Tracing-Last & 0.597 & 0.604 & 0.399 & 0.540&0.449&0.279&0.616 & 0.630 & 0.400 &0.462&0.428&0.279
\\
Rank-Last & \textcolor{red}{0.617} & \textcolor{red}{0.623} & \textcolor{blue}{0.463} &0.534 &\textcolor{red}{0.471}&0.204&\textcolor{blue}{0.663} & \textcolor{blue}{0.671} & \textcolor{blue}{0.587} &0.422&0.452&0.192
\\
\textbf{SWBCE-Last} & \textcolor{blue}{0.620} & \textcolor{blue}{0.625} & \textcolor{red}{0.441} & \textcolor{blue}{0.591} &\textcolor{blue}{0.488}&\textcolor{blue}{0.181}&\textcolor{red}{0.660} & \textcolor{red}{0.670} & \textcolor{red}{0.569}  &\textcolor{blue}{0.532}&\textcolor{red}{0.496}&\textcolor{blue}{0.173}
\\
\hline
\end{tabular}
\end{table}

\begin{table}[htbp]
\renewcommand\arraystretch{0.8}
\centering
\caption{\textbf{Results of BDCN on BIPED2 and BRIND with 1-pixel error tolerance without NMS.} }
\label{BDCN-BIPED2-BRIND}
\small
\begin{tabular}{|@{}c@{}|@{\hspace{0.5mm}}c@{\hspace{1mm}}c@{\hspace{1mm}}c@{\hspace{1mm}}c@{\hspace{1mm}}c@{\hspace{0.5mm}}c@{}|@{\hspace{0.5mm}}c@{\hspace{1mm}}c@{\hspace{1mm}}c@{\hspace{1mm}}c@{\hspace{1mm}}c@{\hspace{0.5mm}}c@{}|}
 \hline
 &\multicolumn{6}{c}{BIPED2}&\multicolumn{6}{c|}{BRIND}
 \\
\hline
    & ODS   & OIS   & AP &SSIM&ER&RMSE & ODS   & OIS   & AP &SSIM&ER&RMSE
    \\
\hline
WBCE & \textcolor{blue}{0.629} & \textcolor{blue}{0.635} & 0.421 &\textcolor{blue}{0.616}&\textcolor{red}{0.481}&0.294& 0.659 & 0.672 & 0.467&\textcolor{blue}{0.523}&0.470&0.294
\\
Tracing & 0.601 & 0.609 & 0.449&0.453&0.430&0.262 & 0.645 & 0.655 & 0.567 &0.397&0.434&0.248
\\
\textbf{SWBCE} & 0.622 & 0.625 & 0.437&\textcolor{red}{0.591}&\textcolor{blue}{0.484}&\textcolor{red}{0.184} & \textcolor{blue}{0.669} & \textcolor{blue}{0.681} & \textcolor{blue}{0.571} &0.507&\textcolor{red}{0.478}&\textcolor{red}{0.176}
\\
Tracing-Last & 0.601 & 0.609 & 0.418&0.520&0.450&0.273 & 0.635 & 0.646 & 0.428 &0.414&0.422&0.260
\\
Rank-Last & \textcolor{red}{0.626} & \textcolor{red}{0.631} & \textcolor{blue}{0.466}&0.464&0.445&0.207 & 0.660 & 0.671 & 0.564 &0.362&0.417&0.200
\\
\textbf{SWBCE-Last} & 0.624 & 0.628 & \textcolor{red}{0.451}&0.574&0.482&\textcolor{blue}{0.183} & \textcolor{red}{0.662} & \textcolor{red}{0.673} & \textcolor{red}{0.569} &\textcolor{red}{0.502}&\textcolor{blue}{0.481}&\textcolor{blue}{0.173}
\\
\hline
\end{tabular}
\end{table}

\begin{table}[htbp]
\renewcommand\arraystretch{0.8}
\centering
\caption{\textbf{Results of Dexi on BIPED2 and BRIND with 1-pixel error tolerance without NMS.}}
\label{Dexi-BIPED2-BRIND}
\small
\begin{tabular}{|@{}c@{}|@{\hspace{0.5mm}}c@{\hspace{1mm}}c@{\hspace{1mm}}c@{\hspace{1mm}}c@{\hspace{1mm}}c@{\hspace{0.5mm}}c@{}|@{\hspace{0.5mm}}c@{\hspace{1mm}}c@{\hspace{1mm}}c@{\hspace{1mm}}c@{\hspace{1mm}}c@{\hspace{0.5mm}}c@{}|}
 \hline
 &\multicolumn{6}{c}{BIPED2}&\multicolumn{6}{c|}{BRIND}
 \\
\hline
    & ODS   & OIS   & AP &SSIM&ER&RMSE & ODS   & OIS   & AP &SSIM&ER&RMSE
    \\
\hline
WBCE & 0.632 & \textcolor{blue}{0.636} & 0.432&0.636&0.499&0.293 & \textcolor{red}{0.666} & 0.672 & 0.478&0.549&0.484&0.291
\\
Tracing & \textcolor{red}{0.651} & \textcolor{red}{0.658} & \textcolor{blue}{0.513}&0.624&\textcolor{blue}{0.554}&0.244 & \textcolor{red}{0.666} & \textcolor{red}{0.675} & \textcolor{blue}{0.565} &\textcolor{red}{0.569}&\textcolor{blue}{0.563}&0.231
\\
\textbf{SWBCE} & 0.643 & 0.648 & 0.472 &0.613&\textcolor{red}{0.524}&\textcolor{red}{0.176}& \textcolor{blue}{0.673} & \textcolor{blue}{0.681} & \textcolor{red}{0.556} &\textcolor{blue}{0.587}&\textcolor{red}{0.549}&\textcolor{red}{0.172}
\\
Tracing-Last & 0.637 & 0.643 & 0.487&0.545&0.499&0.254 & 0.642 & 0.655 & 0.512 &0.503&0.506&0.233
\\
Rank-Last & 0.468 & 0.485 & 0.275 &0.239&0.306&0.198& 0.527 & 0.544 & 0.406 &0.560&0.486&\textcolor{blue}{0.168}
\\
\textbf{SWBCE-Last} & \textcolor{blue}{0.661} & \textcolor{blue}{0.664} & \textcolor{red}{0.504}&0.623&\textcolor{red}{0.528}&\textcolor{blue}{0.170} & 0.663 & 0.672 & 0.549 &0.562&0.524&\textcolor{red}{0.172}
\\
\hline
\end{tabular}
\end{table}

\subsection{Supplementary Figures: Precision-Recall and SWBCE-Epoch Curves}

    \begin{figure}[htbp]
        \centering
        \begin{subfigure}{0.24\columnwidth}
             \includegraphics[width=\linewidth]{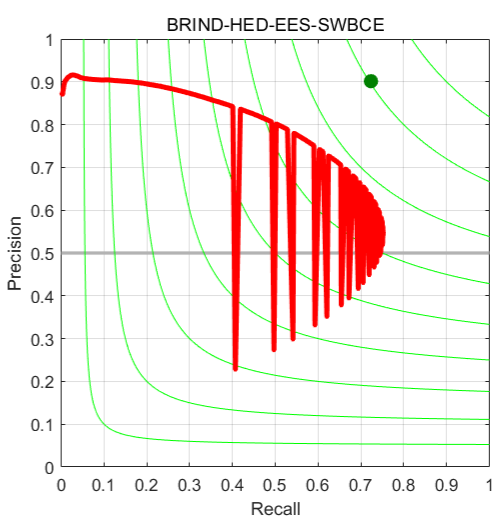}
             \caption{HED-EES}
        \end{subfigure}
        \hfill
        \begin{subfigure}{0.24\columnwidth}
             \includegraphics[width=\linewidth]{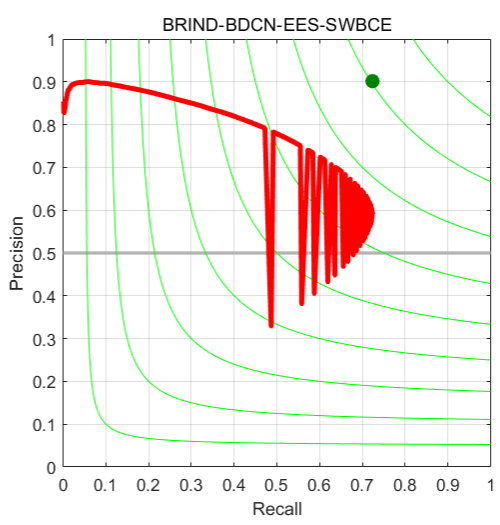}
             \caption{BDCN-EES}
        \end{subfigure}
        \hfill
        \begin{subfigure}{0.24\columnwidth}
             \includegraphics[width=\linewidth]{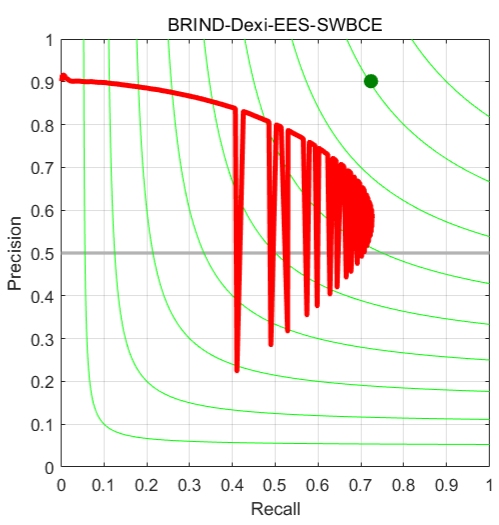}
             \caption{Dexi-EES}
        \end{subfigure}
        \hfill
        \begin{subfigure}{0.24\columnwidth}
             \includegraphics[width=\linewidth]{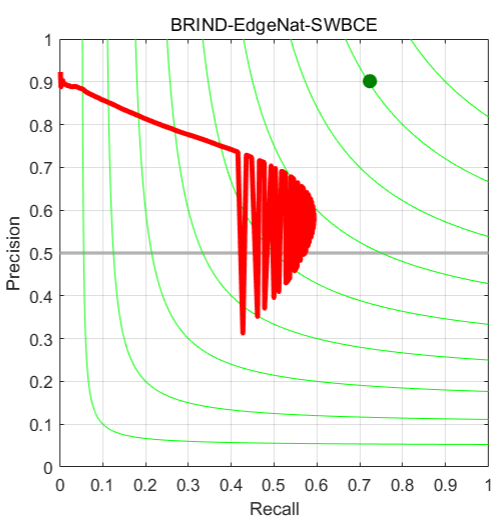}
             \caption{EdgeNat}
        \end{subfigure}
        \caption{\ Precision vs.\ Recall on BRIND with 1-pixel error toleration without NMS, training by SWBCE.}
        \label{fig:BRINDPR}
    \end{figure}

    \begin{figure}[htbp]
        \centering
        \begin{subfigure}{0.24\columnwidth}
             \includegraphics[width=\linewidth]{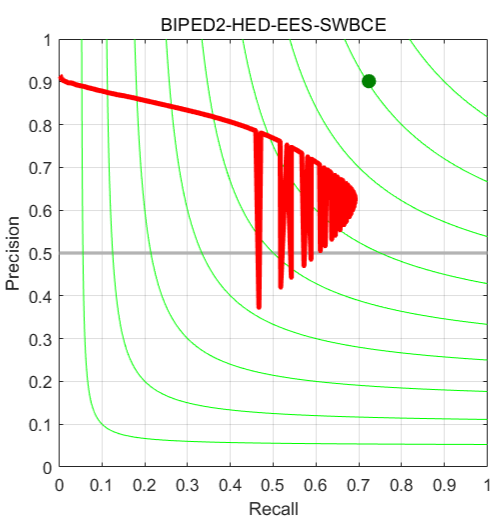}
             \caption{HED-EES}
        \end{subfigure}
        \hfill
        \begin{subfigure}{0.24\columnwidth}
             \includegraphics[width=\linewidth]{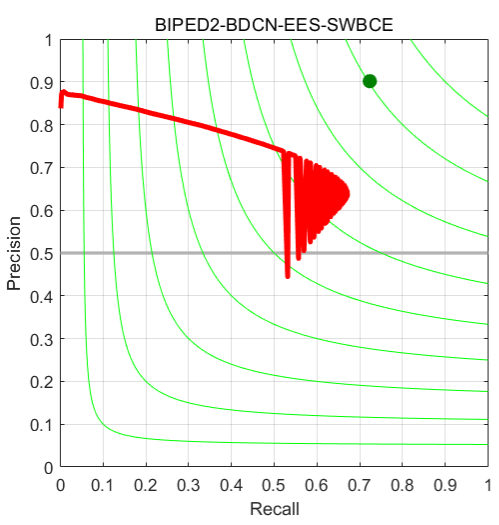}
             \caption{BDCN-EES}
        \end{subfigure}
        \hfill
        \begin{subfigure}{0.24\columnwidth}
             \includegraphics[width=\linewidth]{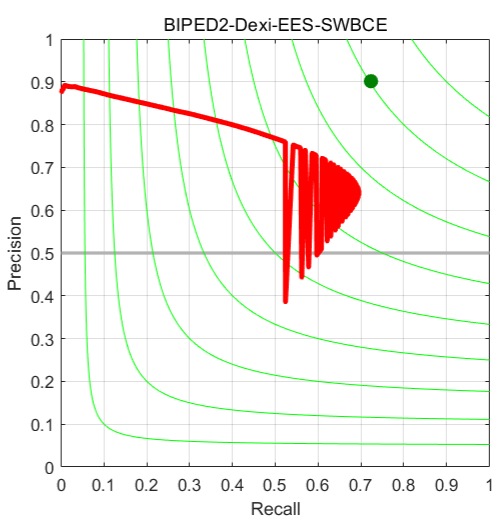}
             \caption{Dexi-EES}
        \end{subfigure}
        \hfill
        \begin{subfigure}{0.24\columnwidth}
             \includegraphics[width=\linewidth]{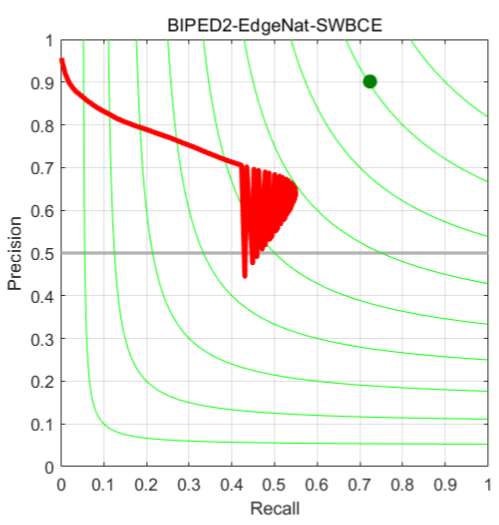}
             \caption{EdgeNat}
        \end{subfigure}
        \caption{\ Precision vs.\ Recall on BIPED2 with 1-pixel error toleration without NMS, training by SWBCE.}
        \label{fig:BIPED2PR}
    \end{figure}

    \begin{figure}[htbp]
        \centering
        \begin{subfigure}{0.24\columnwidth}
             \includegraphics[width=\linewidth]{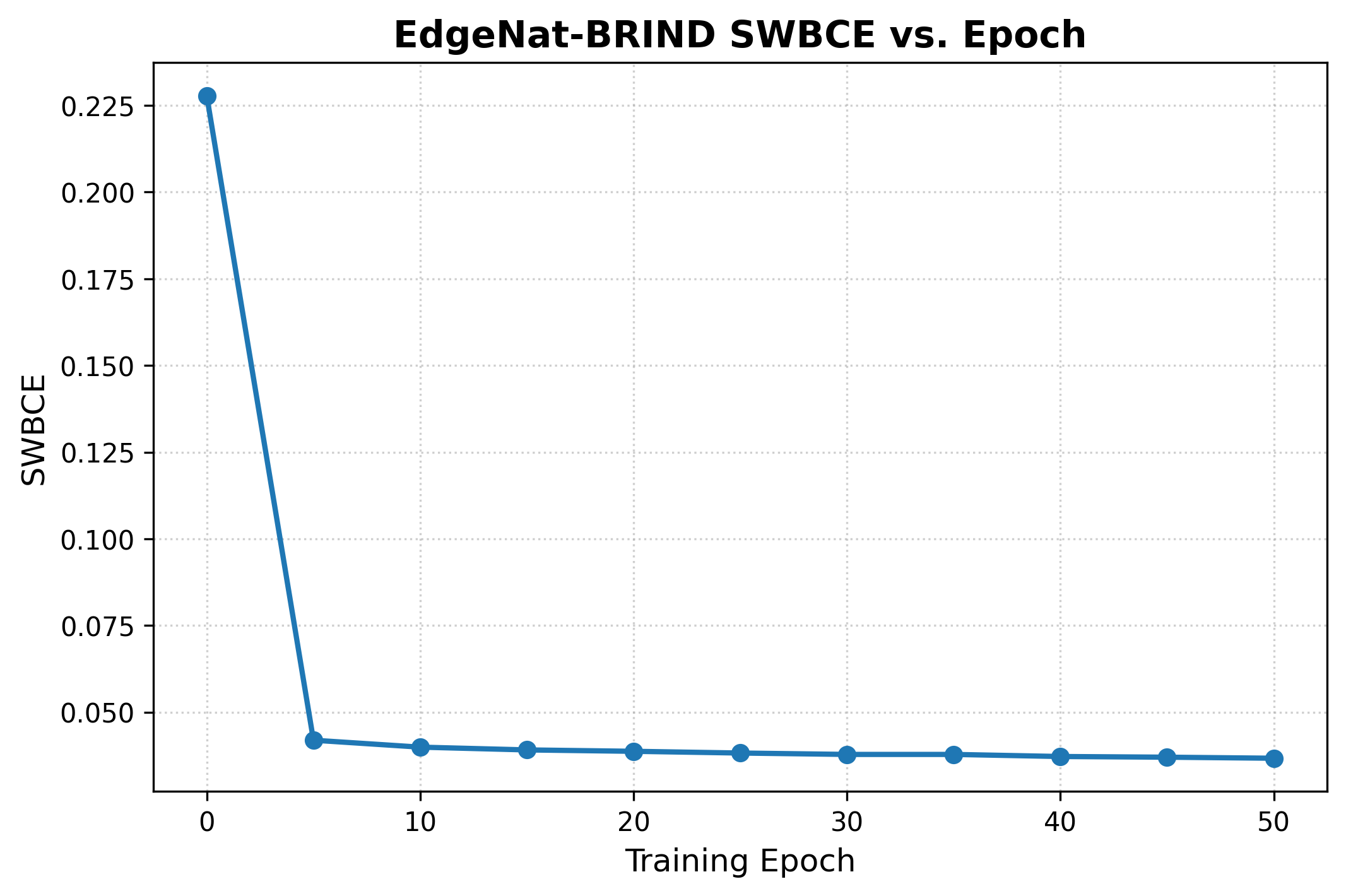}
             \caption{BRIND}
        \end{subfigure}
        \hfill
        \begin{subfigure}{0.24\columnwidth}
             \includegraphics[width=\linewidth]{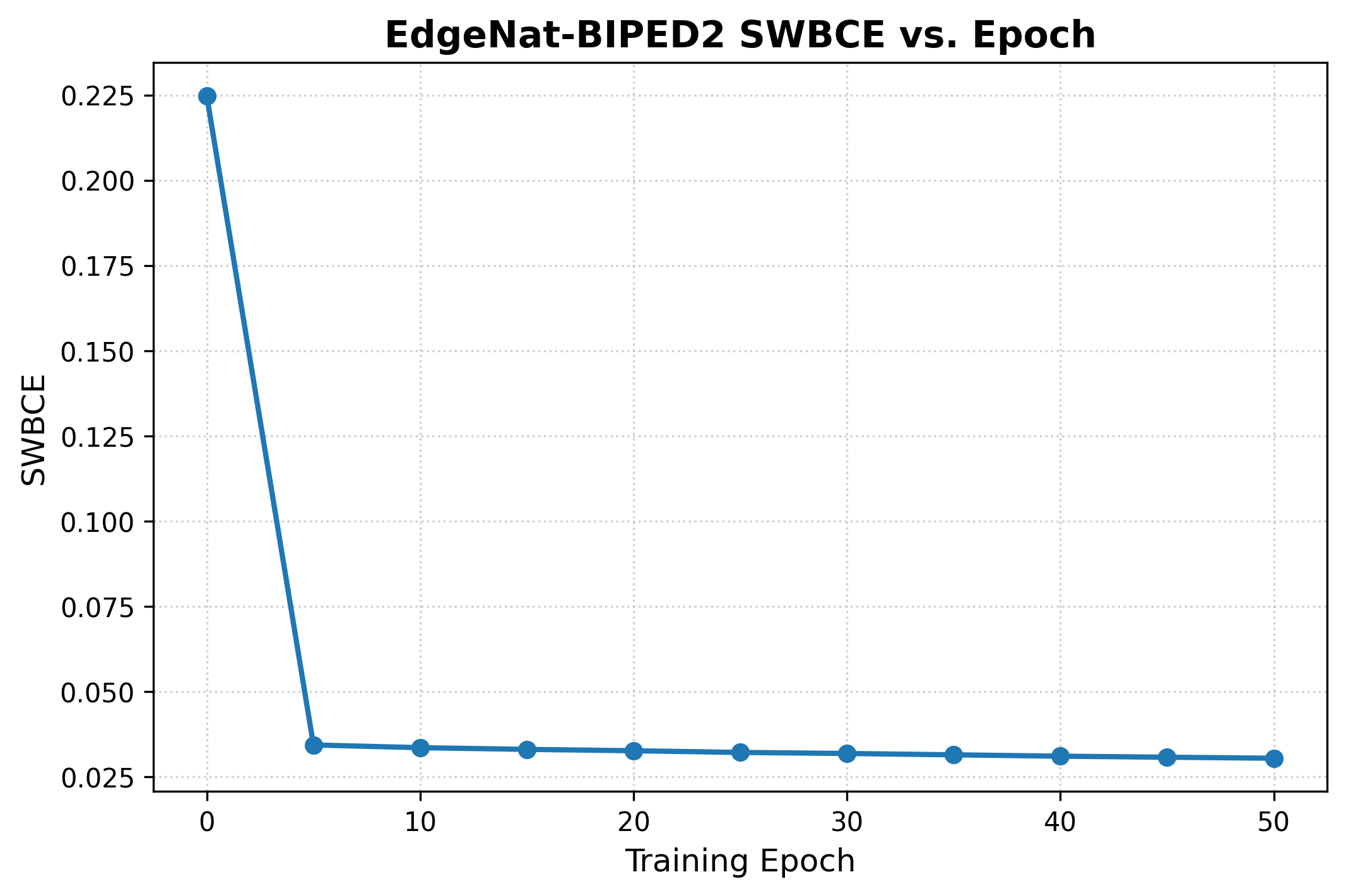}
             \caption{BIPED2}
        \end{subfigure}
        \hfill
        \begin{subfigure}{0.24\columnwidth}
             \includegraphics[width=\linewidth]{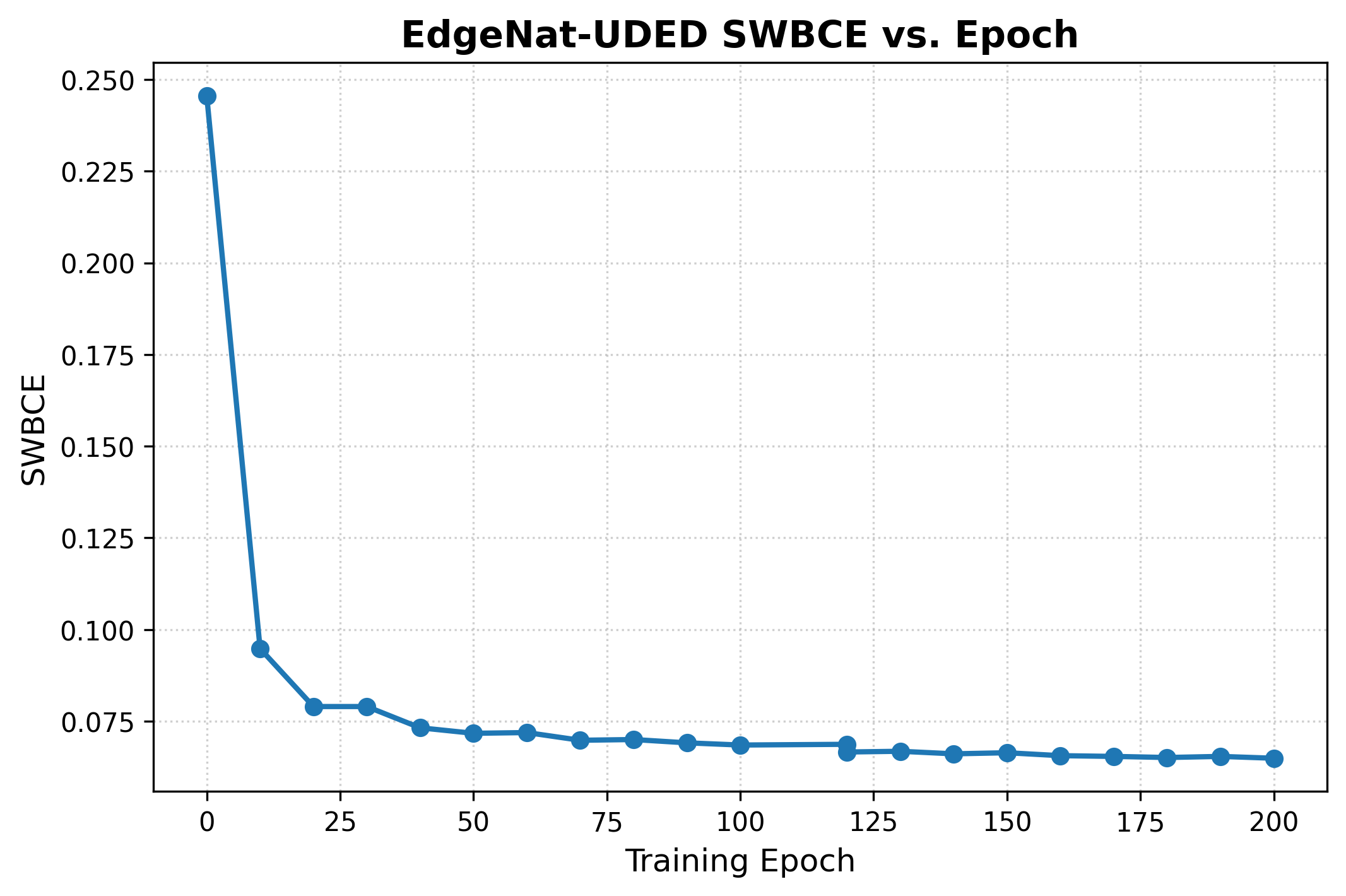}
             \caption{UDED}
        \end{subfigure}
        \hfill
        \begin{subfigure}{0.24\columnwidth}
             \includegraphics[width=\linewidth]{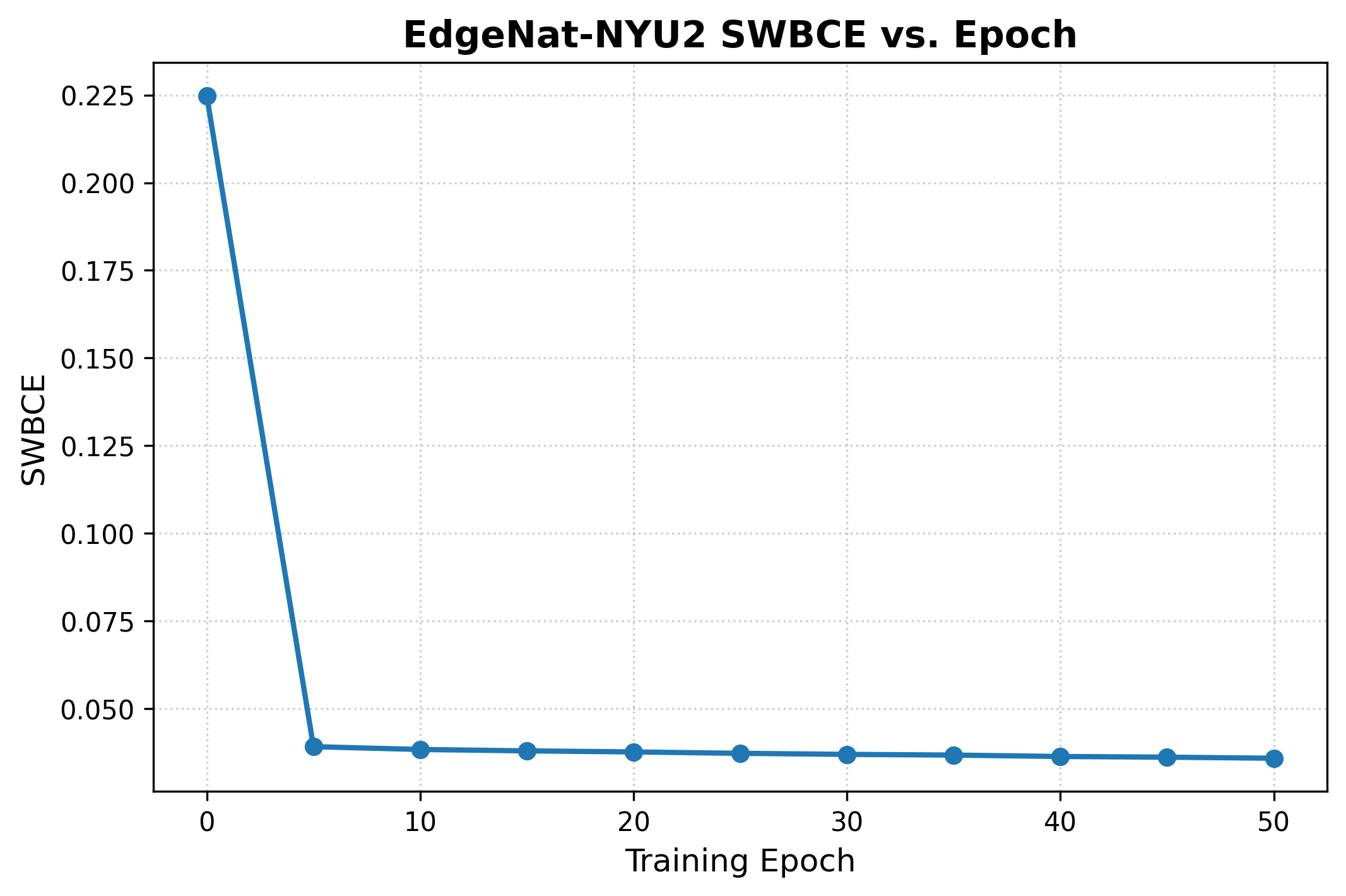}
             \caption{NYU2}
        \end{subfigure}
        \caption{\ SWBCE vs.\ training epochs on EdgeNat. For BRIND, BIPED2, and NYU2, the model converges after 5 epochs, while for UDED, after about 50 epochs.}
        \label{fig:BIPED2PR}
    \end{figure}

\end{document}